\documentclass[lettersize,10pt,twocolumn]{article}
\usepackage[utf8]{inputenc}
\usepackage{amsmath,graphicx}
\usepackage{amsfonts}
\usepackage{amssymb}
\usepackage{algorithm}
\usepackage{algpseudocode}
\usepackage[nice]{nicefrac}
\usepackage{url}
\usepackage{mathtools}
\usepackage{multirow}
\usepackage[table]{xcolor}
\usepackage{tabularx}
\usepackage[textwidth=1.145\textwidth,textheight=1.2\textheight]{geometry}
\usepackage{sectsty}
\usepackage{cleveref}
\usepackage{ulem}
\sectionfont{\normalsize}
\subsectionfont{\normalsize}

\title{\LARGE Temporal Image Forensics: A Review and Critical Evaluation}
\author{\normalsize Robert Jöchl and Andreas Uhl\\\normalsize University of Salzburg, Department of Artificial Intelligence and Human Interfaces, \\\normalsize Salzburg, Austria\\
\ \normalsize \{robert.joechl, andreas.uhl\}@plus.ac.at}
\date{}

\DeclareMathOperator*{\argmax}{argmax} 

\begin{document}

\maketitle

\begin{abstract}
Temporal image forensics is the science of estimating the age of a digital image. Usually, time-dependent traces (age traces) introduced by the image acquisition pipeline are exploited for this purpose. In this review, a comprehensive overview of the field of temporal image forensics based on time-dependent traces from the image acquisition pipeline is given. This includes a detailed insight into the properties of known age traces (i.e., in-field sensor defects and sensor dust) and temporal image forensics techniques. Another key aspect of this work is to highlight the problem of content bias and to illustrate how important eXplainable Artificial Intelligence methods are to verify the reliability of temporal image forensics techniques. Apart from reviewing material presented in previous works, in this review: (i) a new (probably more realistic) forensic setting is proposed; (ii) the main properties (growth rate and spatial distribution) of in-field sensor defects are verified; (iii) it is shown that a method proposed to utilize in-field sensor defects for image age approximation actually exploits other traces (most likely content bias); (iv) the features learned by a neural network dating palmprint images are further investigated; (v) it is shown how easily a neural network can be distracted from learning age traces. For this purpose, previous work is analyzed, re-implemented if required and experiments are conducted.
\end{abstract}

\section{Introduction}
\label{sec:intro}
In the field of temporal image forensics, the main objective is to approximate (estimate) the age of a digital image. Age is typically regarded as a relative age. This age information may help a forensic analyst to establish a chronological sequence between pieces of evidence (i.e., digital images). A chronological sequence of images might, for example, assist in identifying causal relationships between events, which could be relevant in court trials. The most obvious way to determine the age of an image is to check the header information (EXIF). However, the acquisition timestamp stored in the EXIF header may have been lost or manipulated. In general, the information stored in the EXIF header is easy to manipulate and therefore not trustworthy. For this reason, methods that approximate the age of a digital image are required.

A way to approximate the age of an image (or the time of acquisition) is to analyze the image content. For example, the GoogLeNet\cite{GoogleNet} is utilized in \cite{Mueller17a} for image dating based on image content. The authors also introduce the Image Date Estimation in the Wild dataset, where a collection of 1029710 images taken between 1930 and 1999 were downloaded from Flickr. Predicting the acquisition year, the authors report a mean error of less than 8 years. The combination of the estimated age of the entire image with the estimated age of detected objects in the image is proposed in \cite{Ashida21a}. For example, if a car of a specific make and model is detected, the image was taken at least after the first production of that type of vehicle. In \cite{Ginosar15a}, the age of American high school yearbook images captured between 1905 and 2013 is estimated based on human appearance. Another approach that relates human appearance and time is introduced in \cite{Salem16a}. The authors observe that clothing, hair styles, and glasses can be informative features. A method for detection of temporal metadata manipulation based on the image content is proposed in \cite{Padilha22a}. The stored timestamp is verified by checking if the image content, capture time, and geographical location are consistent. However, the use of image content for image age approximation requires the presence of temporally relevant clues (evidence) in the recorded scene. In addition, the maximum temporal resolution of these methods is probably limited.

Another way to approximate the age of an image is to analyze time-dependent traces (e.g., in-field sensor defects) introduced by the image acquisition pipeline (i.e., optical system, sensor, etc.). The idea is that by detecting these time-dependent traces in an image under investigation, the relative age of this image to images from the same device can be estimated. For example, in-field sensor defects (single pixel defects) accumulate over time \cite{Dudas07a}. Thus, by detecting the defects present, the age of an image can be approximated.

This review paper focuses on the field of temporal image forensics based on time-dependent traces (age traces) introduced by the image acquisition pipeline. First (section \ref{sec:setting}), the commonly assumed forensic setting in this context is reflected, the limitations and restrictions are described and a new (probably more realistic) setting is proposed. In section \ref{sec:traces}, the properties of known age traces are described in detail  and compared with observed properties. The observed properties are based on the Paris Lodron University Salzburg Temporal Image Forensics (PLUSTIF) dataset, a dataset we published in a previous work \cite{Joechl24a}. An overview of image age approximation techniques proposed in the field of temporal image forensics is given in section \ref{sec:methods}. Works that have investigated the features learned by Convolutional Neural Network (CNN) in the context of temporal image forensics are covered in section \ref{sec:features}. The problem of content bias is highlighted in section \ref{sec:content_bias} by demonstrating how easily a neural network can be distracted from learning age traces. Existing works in the context of temporal image forensics and eXplainable Artificial Intelligence (XAI) are described in section \ref{sec:explain}. Based on XAI techniques, existing data-driven techniques for image age approximation are analyzed in section \ref{sec:xai_eval}. It is demonstrated that a method claiming to utilize in-field sensor defects exploits other traces (most likely content bias). A discussion is given in section \ref{sec:discussion} and the key insights are summarized in the conclusion (section \ref{sec:con}).

In short, apart from reviewing material presented in earlier works, this review:
\begin{itemize}
 \item proposes a new (probably more realistic) forensic setting.
 \item verifies the main properties (growth rate and spatial distribution) of in-field sensor defects.
 \item highlights the problem of content bias by showing how easily a neural network can be distracted from learning age traces.
 \item shows that a method claiming to utilize in-field sensor defects actually exploits other traces (most likely content bias).
 \item further investigates the features learned of a method dating palmprint images.
 \item provides a discussion.
\end{itemize}

Supplementary material is made available\footnote{https://wavelab.at/sources/Joechl25b/}. This material includes the detailed results of the detected in-field sensor defects in the PLUSTIF dataset, the protocol for creating the standard scene dataset, and the complete results of all experiments conducted for this work.

\section{Defining the Assumed Forensic Setting}
\label{sec:setting}
\begin{figure}[!t]
	\centering
	\includegraphics[width=0.98\linewidth]{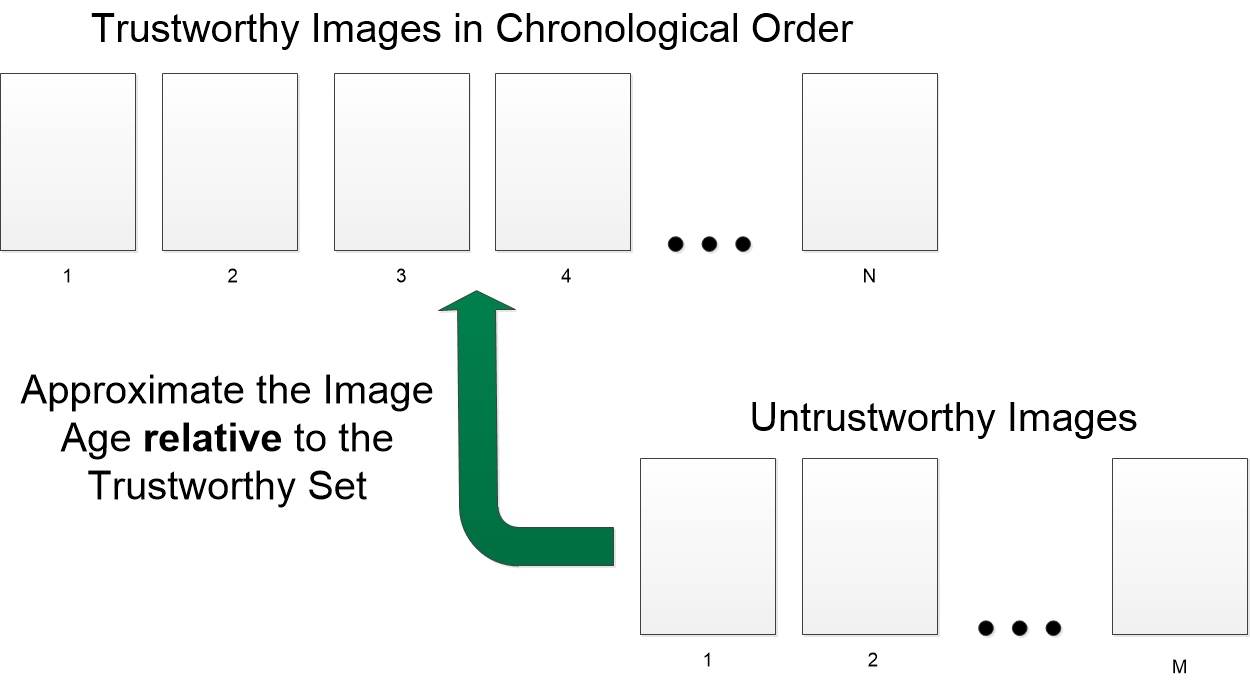}
	\caption{Traditional temporal image forensics investigation setting.}
	\label{fig:traditional_setting}
\end{figure}
In the typical temporal image forensics investigation setting \cite{Fridrich11a,Joechl20a,Ahmed21a,Ahmed20b} (as considered in this work and depicted in Figure \ref{fig:traditional_setting}), it is assumed that a forensics analyst is provided with a set of chronologically ordered (trusted) images and a second set of (untrustworthy) images from the same device with unknown acquisition times. The main objective is to approximate the age of images from the untrustworthy set relative to images from the trusted set. For this purpose, age traces (left during the image acquisition pipeline) are identified and estimated based on images from the trusted set. By detecting the corresponding traces in an image under investigation, a temporal assignment can be established. In this setting, image age approximation can be viewed as a multi-class classification problem, where the classes are defined by the temporal resolution of the considered age traces and the available trustworthy images.

This traditional setting is subject to certain limitations and restrictions: (i) the existence and availability of a trusted set of chronologically ordered images (from the same device) is a rather strong assumption, (ii) developed image age approximation techniques aim to establish relative age information, but not absolute age, and (iii) the eventual availability of the device used to acquire the evidence in question (i.e., the digital image) is not at all exploited.

In response to these observations, a new temporal investigation setting is suggested. In this new setting, it is assumed that a forensics analyst is provided with a set of not trustworthy images of unknown age and the imager with which these images were taken. The availability of the imager could be used to capture calibration images in order to detect age traces that would facilitate a temporal ordering of the images under investigation. Assuming a known manufacturing time of the imager, another interesting field of research would be the estimation of the temporal evolution of the detected age traces, which could enable to estimate the absolute age of an image.

However, the presence of age traces is at the heart of any method to approximate the age of an image in any assumed setting.

\section{Age Traces}
\label{sec:traces}
All components of the image acquisition pipeline (e.g., the optical system, the sensor, etc.) leave certain traces in a digital image. If such traces are not time invariant, they can be considered as age traces and used to approximate the image age. The best known and most commonly used traces in this context are in-field sensor defects.

\subsection{In-Field Sensor Defects}
\begin{figure}[!t]
    \centering
    \begin{minipage}[b]{0.48\textwidth}
		\centering
		\includegraphics[width=0.88\textwidth]{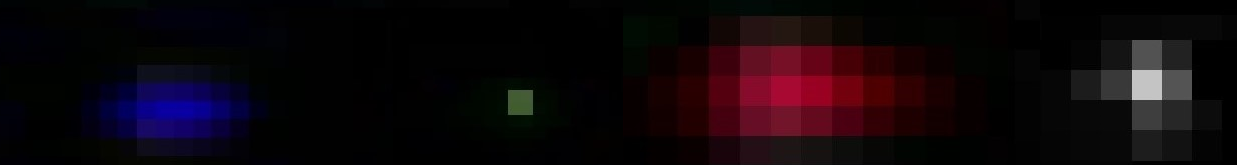}
		\centerline{(a)}\medskip
	\end{minipage}
    \begin{minipage}[b]{0.48\textwidth}
		\centering
		\includegraphics[width=0.88\textwidth]{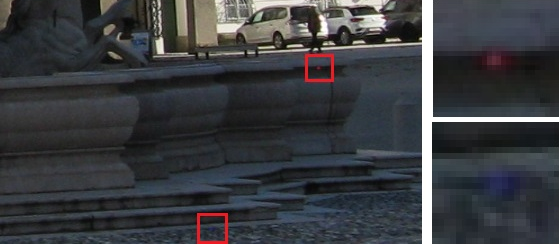}
		\centerline{(b)}\medskip
	\end{minipage}
  \caption{(a) Defects extracted from a Dark Field Image; (b) Defects present in a regular scene image.}
  \label{fig:defect_example}
\end{figure}
Just as human beings age, so do image sensors. In the case of an image sensor, the development of single pixel defects is a sign of such aging. These defects are termed in-field sensor defects because they develop after the manufacturing process. In principle, the defect is isolated to a single pixel only. In \cite{Leung07a}, it is statistically estimated that a defect affects less than 2.3\% of a pixel's area. However, the pixel defects are distorted by several preprocessing steps such as demosaicing and image compression. For example, when a bilinear demosaicing algorithm is applied, the defect spreads to all neighboring pixels \cite{Chapman19a}. An example of a defect extracted from a Dark Field Image (DFI) is given in Figure \ref{fig:defect_example} (a). DFIs are calibration images where the camera's shutter is closed, thereby setting the incident light to zero. In \cite{Chapman19a}, it is shown that when two defective pixels occur within a $5 \times 5$ pixel block, the defects can spread to a $16 \times 16$ block due to demosaicing and JPEG compression. An example of defects present in a regular scene image is given in Figure \ref{fig:defect_example} (b).

\subsubsection{Temporal Growth}
In \cite{Dudas07a}, the characteristics of in-field sensor defects are studied by analyzing calibration images (i.e., illuminating the sensor with uniform fields) taken at different time-slots. It is empirically shown that these defects develop instantaneously (i.e., switch from their good to defective state between two consecutive captured images) and remain defective throughout the lifetime of the sensor. The number of defects increase linearly with time and a growth rate of 0.035 defects/month/megapixel or 0.082 defects/1000 images/megapixel is observed.

By applying statistical analysis (i.e., Chi-Square test) it is shown in \cite{Leung07a} that the inter-defect time follows an exponential distribution with constant rate, i.e., defects develop continuously and independently. A growth rate of 0.22 defects/month or 0.52 defects/1000 images is estimated. A difference in defect growth rates between sensor types (i.e., CCD and CMOS (APS)) is first discovered in \cite{Leung08a}. The authors report 2.2 defects/year for APS sensors and 5.2 defects/year for CCD sensors. So far, the evaluations in \cite{Leung07a,Leung08a} have been conducted with a fixed ISO setting (i.e., ISO 400). Since the ISO gain is applied to all pixels (whether good or defective), the defect parameters are also amplified. This leads to defects being more visible at higher ISO settings. In \cite{Leung09b}, the noticeable defects increased from 34 to 250 for ISO settings 100 and 1600 respectively.

Besides the sensor type, the pixel size also has a significant influence on the defect growth rate. The influence of pixel size is first observed in \cite{Leung10a} based on an analysis of DSLRs and cellphone cameras. The high defect rate observed in cellphone cameras suggests that the small pixel size can lead to more defects. Chapman et al. further investigate the influence of pixel size and propose an empirical formula relating defect density to pixel size and ISO setting in \cite{Chapman13a}. In principle, the defect density increases in a power law as the pixel size shrinks, i.e.,
\begin{equation}
 D = A S^B ISO^C,
 \label{eq:defect_rate}
\end{equation}
where $D$ represents the defect density (defects/year/mm$^2$), A denotes the number of defects/year/mm$^2$ when the pixel size is 1 $\mu$m, $S$ is the pixel size, and $B$ and $C$ are constants. Depending on the sensor type, the empirically determined values are: (CCD) $A=10^{-1.849}$, $B=-2.25$ and $C=0.687$; (APS \cite{Chapman24a}) $A=10^{-0.98}$, $B=-3.03$ and $C=0.506$. As can be seen, the defect density grows at a much higher rate with an APS sensor than with a CCD sensor. The authors report that with a pixel size of about 2 microns and ISO 400, the defect density between APS and CCD sensors should be about equal.

\begin{table}
\begin{center}
\caption{Overview of the imagers used to create the PLUSTIF dataset.}
\label{tab:PLUSTIF_imagers}
\small
\begin{tabular}{ l | l | c | c }
ID & Model & Sensor & Pixel Size \\
\hline \hline
Pc01 & Canon PS A720IS & CCD & 1.76$\mu$m \\ \hline
Pc02 & Canon EOS 70D   & CMOS & 4.01$\mu$m\\ \hline
Pf01 & Fujifilm X100V  & CMOS & 3.75$\mu$m\\ \hline
Pk01 & KonicaMinolta Dimage Z5 & CCD & 2.22$\mu$m \\ \hline
Pn01 & Nikon E7600 & CCD & 2.34$\mu$m \\ \hline
Pp01 & Pentax K5  & CMOS & 4.68$\mu$m \\ \hline
Pp02 & Pentax K5II & CMOS & 4.68$\mu$m \\ \hline
Ps01 & Sony DSC-P200 & CCD & 2.34$\mu$m\\ \hline \hline
\end{tabular}
\end{center}
\end{table}

\begin{figure*}[!t]
	\centering
	\includegraphics[width=0.98\linewidth]{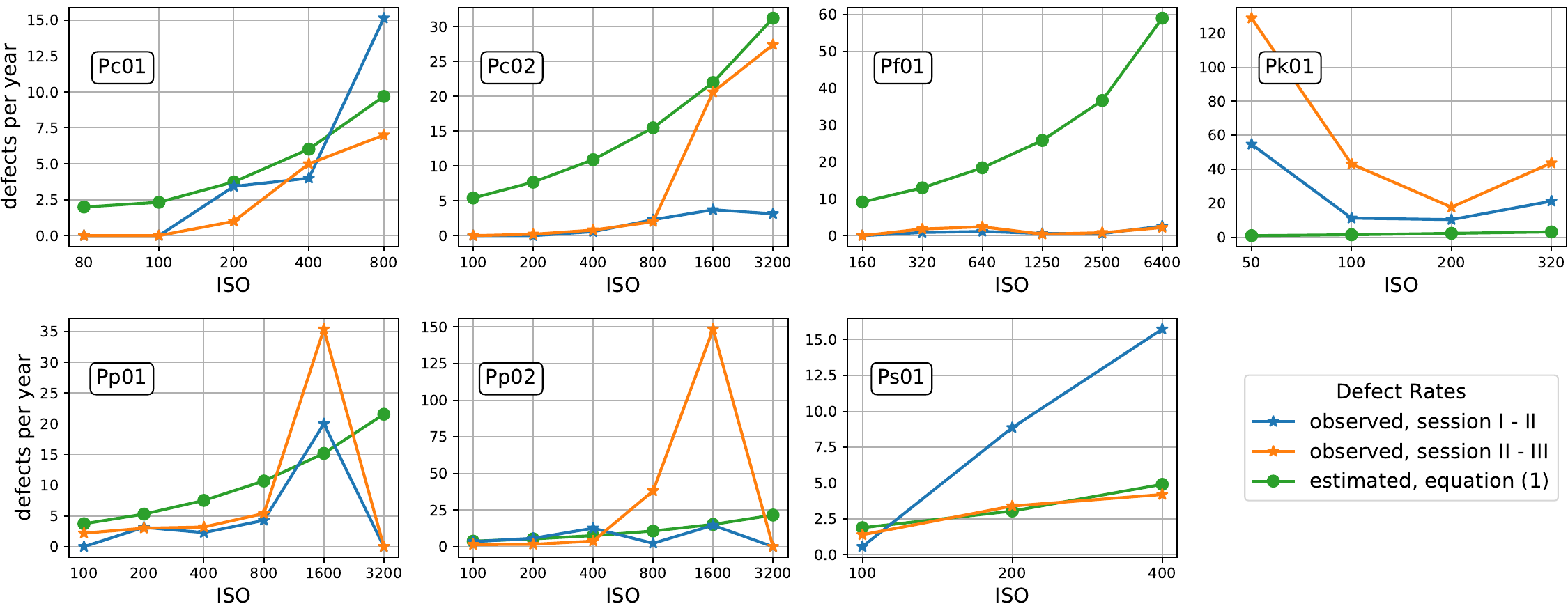}
	\caption{The defect growth rates observed with the PLUSTIF dataset compared to the rates estimated by equation (\ref{eq:defect_rate}).}
	\label{fig:def_dev_rates}
\end{figure*}

In summary, in-field sensor defects develop linearly over time, whereby the growth rate depends on the sensor type and pixel size.

Recently, we proposed the PLUSTIF dataset, in which images from three different time-slots (acquisition sessions) from eight different imagers (see Table \ref{tab:PLUSTIF_imagers}) are available. There is a time difference of around seven months between the first and second acquisition session and around ten months between the second and third acquisition session. The observed defect growth rates are illustrated in Figure \ref{fig:def_dev_rates} and compared with the estimated growth rates based on equation (\ref{eq:defect_rate}). The reported defect rates are based on the in-field sensor defects detected by applying a threshold (i.e., $t=14$) to the captured DFIs. In the PLUSTIF dataset, DFIs with different exposure times are available for each ISO setting. Thus, the average defect growth rate (across all exposure time settings) is illustrated in Figure \ref{fig:def_dev_rates}. The in-field sensor defect developed rates observed only partially align with the estimated rates based on equation (\ref{eq:defect_rate}). Obviously, the observed defect developed rates depend on the exposure time and the applied threshold value. A completely uncharacteristic behavior can be observed with Pk01, where the highest defect growth rates occur at the minimum ISO setting (i.e., $ISO=50$). Both Pentax cameras (i.e., Pp01 and Pp02) exhibit a defect growth rate of 0 at ISO 3200. This is most likely due to some defect suppression technique automatically applied at high ISO settings. The detailed results, i.e., the detected defects for all ISO and exposure time combinations, can be found in the supplementary material. With the Pn01, only DFIs with ISO 200 are available. The observed defect growth rates are 14 and 9 between session I-II and session II-III, respectively. A defect growth rate of 3 is estimated based on the equation (\ref{eq:defect_rate}).

\subsubsection{Spatial Distribution}
In-field sensor defects are randomly distributed across the image sensor. A statistical analysis (i.e., Chi-Square test) of the inter-defect distance distribution is performed in \cite{Leung07a}. The authors report a uniform spatial distribution with no significant bias towards short or long distances (i.e., no clustering). This result is also confirmed in \cite{Leung09a, Leung10a}. In \cite{Chapman19a}, the number of defective pixels required to achieve a given probability that two defects occur within a $5 \times 5$ pixel block is estimated. Considering a 20 megapixel DSLR camera, 128 defects generate a 4\% probability of two such closely spaced defects.

\begin{figure}[!t]
	\centering
	\includegraphics[width=0.68\linewidth]{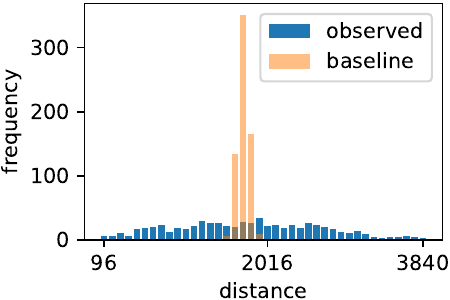}
	\caption{The observed inter-defect distance distribution (Ps01) compared with the distribution that results when the same number of defects are distributed uniformly across the sensor (baseline).}
	\label{fig:inter_defect_dist}
\end{figure}

In Figure \ref{fig:inter_defect_dist}, the observed inter-defect distance distribution for the Ps01 is compared with the distribution that results when the same number of defects is distributed uniformly across the sensor. As can be seen, no bias towards shorter or longer distances could be observed (i.e., no clustering). Contrary to reports in the literature, however, a truly uniform distribution of defects could not be confirmed. Similar results are obtained for all other imagers.

\subsubsection{Defect Cause}
The described continuous development of new defects, the distribution across the sensor (i.e., no clustering) and a defect size smaller than the pixel size indicate constant stress and a random source as a defect cause (e.g., cosmic radiation). Defects would appear in clumps (cluster) if environmental stress (material degradation) would be the reason. In \cite{Theuwissen07a}, the influence of terrestrial cosmic rays on the reliability of CCD sensors is investigated. Terrestrial cosmic rays are the result when primary cosmic rays (i.e., emitted by the sun and/or created in space) hit the earth's atmosphere \cite{Ziegler96a}. The energy and density of terrestrial cosmic rays depends on the altitude and latitude. For this reason, in \cite{Theuwissen07a} sensors are stored on a shelf (40m above sea level), in a laboratory at Jungfraujoch (3450m above sea level), in an underground laboratory (250m below sea level, where all incoming cosmic rays are absorbed) and flown around in an airplane. The density of neutrons in the cosmic ray flux is about a factor of 100 higher during airplane trip and a factor of 10 higher on the Jungfraujoch than at sea level. These factors correlate very well with the observed increase in defects and indicate that neutrons from cosmic rays are the main defect source. In particular, neutrons can create a displacement damage (vacancy and interstitial) in the silicon bulk of a pixel. Neutrons as a defect source implies that the creation of new defects is independent of technology, architecture, sensor type, and sensor vendor \cite{Theuwissen07a}.

A factor 10 to 1000 improvement for large-amplitude defects is observed when storing the sensor in an underground laboratory. A higher improvement was expected because the laboratory is absolutely free of cosmic rays. Most surprising, however, is that almost no improvement is detected for defects with small-amplitudes. This could indicate that more than just cosmic rays probably influence the development of in-field sensor defects.

The effect of temperature when storing an image sensor is evaluated in \cite{Theuwissen08a}. For this purpose, the sensors are stored for several weeks at $60^{\circ}$C, $85^{\circ}$C, $110^{\circ}$C, $150^{\circ}$C and $180^{\circ}$C, respectively. Storage of the sensors at $180^{\circ}$C completely suppresses the development of large-amplitude defects. However, the development of small-amplitude defects increases compared to room temperature. Based on this observation, the author suggests storage at $85^{\circ}$C, where the generation of large-amplitude defects is considerably reduced and the development of small-amplitude defects is similar to room temperature. In addition, an annealing effect is observed when the sensors are stored at higher temperatures. For example, after storing the sensors 24h at $110^{\circ}$C close to 100\% of the defects are annealed. A possible explanation for the observed annealing could be that the mobility of the vacancies and the interstitial silicon atoms increases during storage at higher temperatures and this leads to recombination.

\subsubsection{Defect Types}
In principle, there are several types of defects. A fully stuck pixel has completely lost sensitivity to incoming light and only outputs a constant value. Partially stuck pixels are still sensitive to incoming light, but with a fixed offset. Depending on this offset, the dynamic range of the pixel is more or less reduced. Pixels with abnormal sensitivity have non-unity sensitivity. Nevertheless, the most common defect type is a hot pixel. In fact, there are no fully stuck and abnormal sensitivity pixels found in \cite{Dudas06a,Dudas07a,Leung08b,Leung10a,Chapman13a}. All partially stuck pixels found are also hot pixels. Thus, the defect type partially stuck hot pixel is introduced in \cite{Leung08b}. Among the observed hot pixels in \cite{Leung10a} and \cite{Chapman13a}, 78\% and 56\% are partially stuck hot pixels, respectively.

\begin{figure}[!t]
\centering
\includegraphics[width=0.46\textwidth]{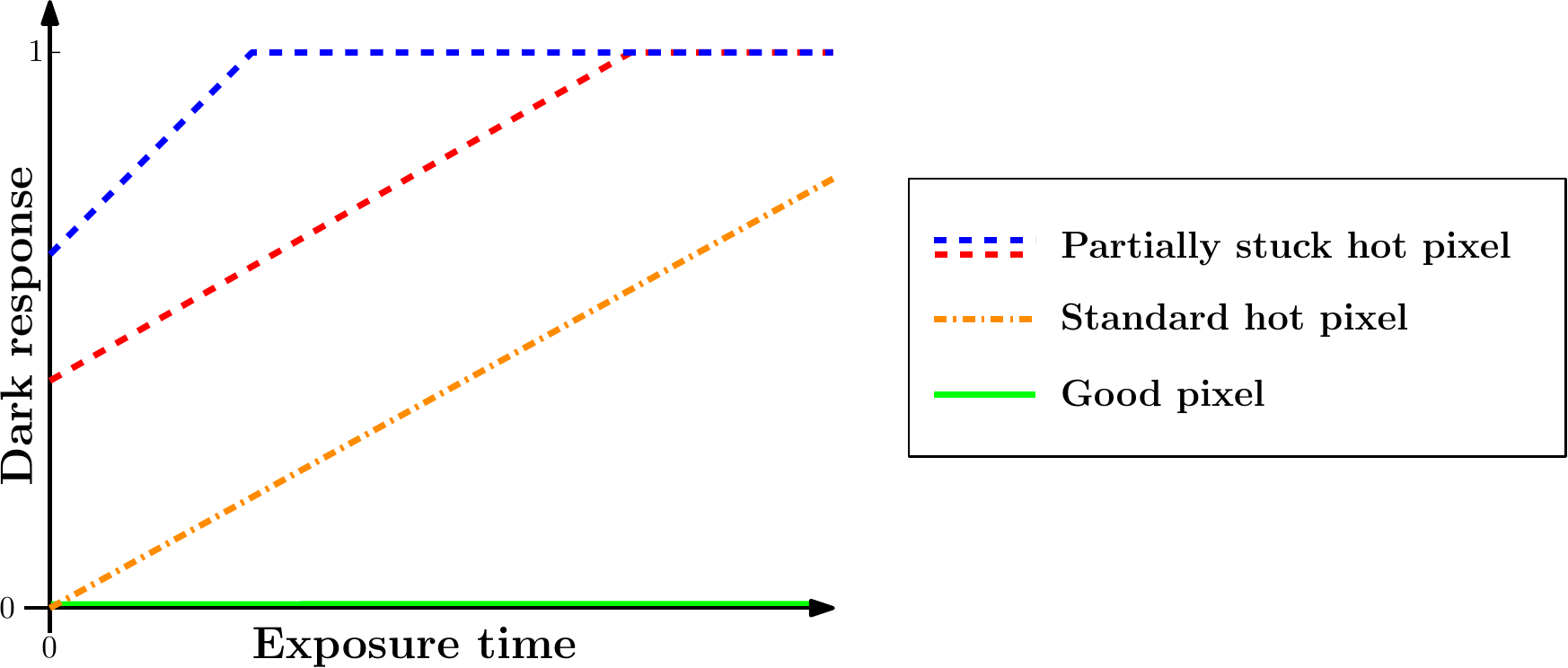}
\caption{Sketch of the dark response of hot pixel types compared to a good pixel, based on the observations in \cite{Leung09b}.}
\label{fig:dark_response}
\end{figure}

As described in \cite{Leung09a}, hot pixels have an illumination independent component, the dark current, which increases linearly with exposure time (see Figure \ref{fig:dark_response}). Depending on the offset, a hot pixel can be fully saturated even at short exposure times. Furthermore, the magnitude of the defect parameters of a hot pixel increase with ISO amplification. At ISO 1600 2-3\% of the defective pixels detected in \cite{Leung09b} are fully saturated. In such a case, a hot pixel is degraded to a fully stuck pixel. A statistical evaluation of hot pixel parameters shows that stuck high pixels are in fact hot pixels with a strong offset \cite{Chapman17a}.

\subsubsection{Defect Model}
The response of a defective pixel to an incident illumination $I$ is modeled in \cite{Fridrich11a} as follows:
\begin{equation}
 Y = I + IK + \tau D + c + \Theta,
 \label{eq:pixel_model}
\end{equation}
where $K$ denotes the Photo-Response Non-Uniformity (PRNU) and the dark current is represented by $D$. Since the dark current is dependent on the exposure time, ISO setting and temperature, these factors are combined in $\tau$. A potential fixed offset is denoted by $c$ and all other noise sources are combined in $\Theta$. The PRNU is caused by the impurities in silicon wafers and manufacturing imperfections that affect the light sensitivity of each individual pixel and cause a fixed noise pattern \cite{Dirik08a}. However, the PRNU should not alter over time. With a good pixel, $D$ and $c$ are equal to zero.

A slightly different model for the response of a pixel to an incident illumination $I$ is given in \cite{Chapman19a} by:
\begin{equation}
 Y = m * (I T_{e} + D T_{e} + c).
 \label{eq:pixel_model2}
\end{equation}
$m$ denotes the ISO amplification and $T_{e}$ the exposure time. This model neglects the additive noise and the PRNU. In addition, the ISO amplification $m$ is not only applied to the dark current $D$ (as in equation (\ref{eq:pixel_model})), but to the entire pixel response.\\

\begin{figure*}[htb!]
    \centering
    \begin{minipage}[b]{0.16\textwidth}
		\centering
		\includegraphics[width=0.98\textwidth]{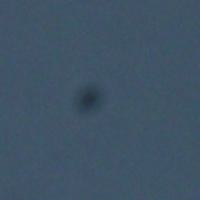}
		\centerline{F/32}\medskip
	\end{minipage}
	\begin{minipage}[b]{0.16\textwidth}
		\centering
		\includegraphics[width=0.98\textwidth]{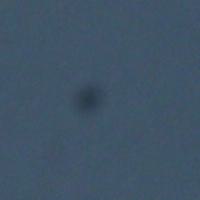}
		\centerline{F/28}\medskip
	\end{minipage}
	\begin{minipage}[b]{0.16\textwidth}
		\centering
		\includegraphics[width=0.98\textwidth]{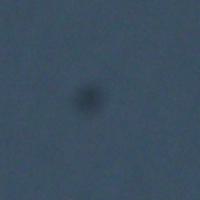}
		\centerline{F/26}\medskip
	\end{minipage}
	\begin{minipage}[b]{0.16\textwidth}
		\centering
		\includegraphics[width=0.98\textwidth]{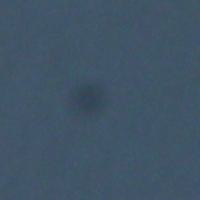}
		\centerline{F/23}\medskip
	\end{minipage}
	\begin{minipage}[b]{0.16\textwidth}
		\centering
		\includegraphics[width=0.98\textwidth]{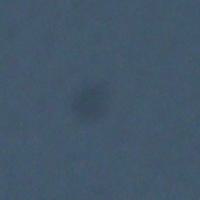}
		\centerline{F/20}\medskip
	\end{minipage}
	\begin{minipage}[b]{0.16\textwidth}
		\centering
		\includegraphics[width=0.98\textwidth]{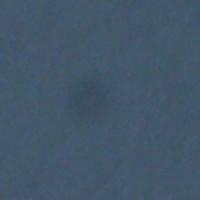}
		\centerline{F/18}\medskip
	\end{minipage}

  \caption{The same dust spot in images captured with different f-numbers.}
  \label{fig:sensor_dust}
\end{figure*}

As in-field sensor defects accumulate over time, they are precious age traces. Although, the image quality degrades as the number of defects increases. For this reason, better cameras in particular use concealment techniques to hide defective pixels (e.g., by interpolating defective pixel values with neighboring pixels). In \cite{Kauba21a}, we assess the impact of sensor ageing on fingerprint capturing devices (including optical, capacitive and thermal ones). The fingerprint sensors traveled on an air-plane for 127 days and samples were taken before and after the trip. When analyzing the samples, no in-field sensor defects could be found. This indicates that the evaluated fingerprint sensors either do not develop any in-field defects or they are able to suppress these single pixel defects.

\subsection{Sensor Dust}
\label{sec:sensor_dust}
One advantage of a DSLR camera is that it allows to work with multiple lenses. However, during the process of mounting/unmounting the interchangeable lens, very small dust particles are attracted to the camera and settle on the protective element in front of the sensor. These small dust particles cause a shadow that can be seen as spots on an image and is referred to as a dust pattern \cite{Dirik08a}. The size of the aperture $A$ (f-number), the focal length $f$ and the distance $t$ between the dust particle and the image sensor are the most important factors that determine the shape and darkness of the dust spot. In particular, the diameter $S$ of the sensor dust spot can be obtained by:
\begin{equation}
    S = D \frac{f}{f-t} + A \frac{t}{f-t},
\end{equation}
where $D$ denotes the diameter of the dust particle \cite{Dirik08a}.

In Figure \ref{fig:sensor_dust} the same dust spot for different f-numbers (aperture sizes) is illustrated. At small apertures (high f-number) a dark and strong dust spot is visible. This is because the incoming light can be considered as a pinpoint source (narrow light cone) that can mostly be absorbed by the dust particle. A small f-number, in contrast, causes wider light cones, where the light rays can pass around the dust particle, resulting in broader and blurred dust spots. In other words, the f-number affects the intensity and radius of the dust spot \cite{Dirik08a}.

The position of the dust particle on the image sensor is stable. However, the position where the dust spot appears in an image is dependent on the focal length. As shown in \cite{Dirik08a}, the magnitude of the shift is dependent on the actual position of the dust particle. The farther away the dust particle is from the image origin, the greater the shift when the focal length is changed.

If the sensor is not cleaned, sensor dust is persistent and accumulates over time \cite{Dirik08a}. Based on these properties, sensor dust can be considered as an age trace that can be used for image age approximation. Although, to the best of our knowledge, there is currently no work that utilizes sensor dust for image age approximation. In \cite{Dirik08a}, sensor dust is exploited for camera identification.\\

In-field sensor defects and sensor dust are two examples of age traces. Nevertheless, each component of the image processing pipeline could introduce time-dependent traces that might be exploited for image age approximation. Since certain cameras conceal in-field sensor defects and images from cell phone cameras in particular are increasingly being post-processed, it is important to investigate new (unknown) age traces.

\section{Image Age Approximation Methods}
\label{sec:methods}
The following methods exploit such age traces (primarily in-field sensor defects) to approximate the age of a digital image.

\subsection{Device Temporal Forensics: An Information Theoretic Approach}
\label{sec:mao}
A two-stage framework to address the problem of device temporal forensics is proposed in \cite{Mao09a}. The first step is to estimate the device parameters $P$ using the device outputs. Then, the temporal evolution of the estimates is modeled in a second stage. This implies that only device parameters that are not time-invariant (e.g., reflecting systematic device degradation) are of interest. The authors assume that the temporal evolution of the parameter can be modeled as an independent increment process, i.e., forms a Markov process. Considering $L$ monotonically increasing time intervals $t_1 < t_2 \dots < t_L$ and applying the data processing inequality, then
\begin{equation}
	I(P(t_i);P(t_j)) \geq I(P(t_i);P(t_k)), \quad \forall i < j < k,
\end{equation}
where $I(.)$ represents the mutual information. Hence, the estimation of the mutual information between estimated device parameters can be used for temporal ordering. The authors show that if the device parameters $P(t)$ form a vector of independent, identically distributed i.i.d. Gaussian processes with identical statistics (i.e., identical mean and autocovariance functions), the mutual information between two device parameters can be estimated by their correlation coefficient.

A realization of the framework is demonstrated by the temporal ordering of images based on the PRNU as the considered device parameter. For this purpose, the images of a given imager are clustered according to their acquisition month, and the PRNU for each cluster is estimated. Then, the correlation coefficients between the estimated PRNU of all clusters are computed. In total, this is done for images of three different devices. Based on the proposed framework, a correct chronological image cluster ordering is possible.

The temporal classification, denoted as Individual Image Placement within an ordered cluster set (IIP), is done in a next step. For this purpose, the PRNU of an image under investigation is estimated. The mutual information (correlation coefficient) of the estimated PRNU with the PRNU estimates of each cluster is calculated. Then, the cluster whose PRNU has the highest mutual information is selected. The authors report correct assignment of test images of 85.3\%, 90.59\% and 96.46\%. However, since the PRNU should not change over time, other time-dependent traces (e.g., in-field sensor defects) are likely implicitly involved in the PRNU estimation.

\subsection{Determining Approximate Age of Digital Images Using Sensor Defects}
\label{sec:fridrich}
\begin{figure}[!t]
\centering
\includegraphics[width=0.48\textwidth]{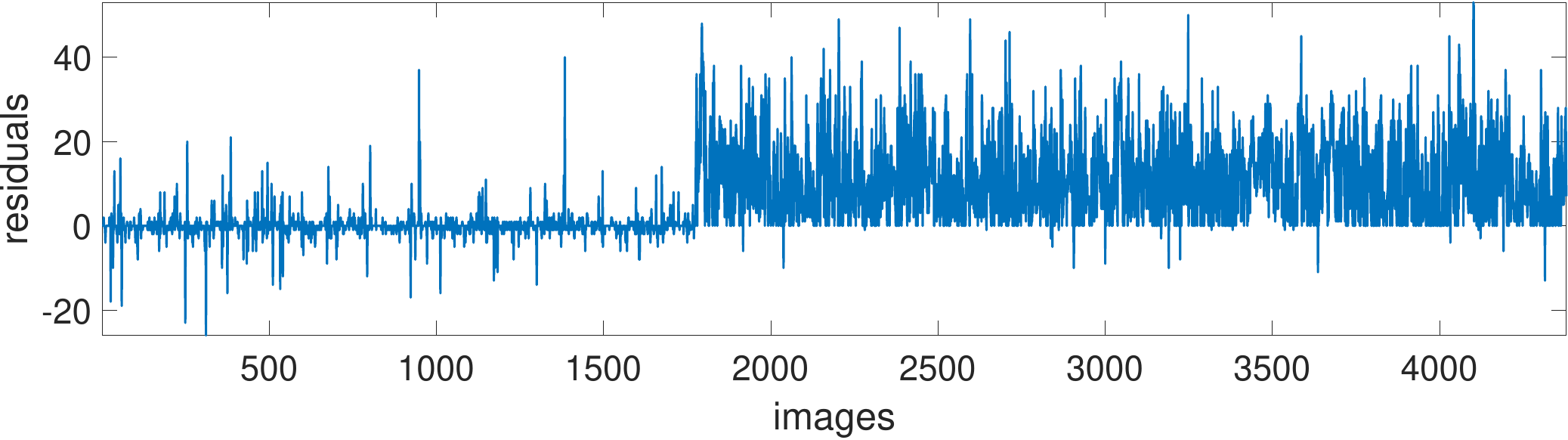}
\caption{Median filter residuals of a defective pixel over chronologically ordered images, as illustrated in \cite{Joechl20a}. The onset of the defect at about index 1770 is clearly visible.}
\label{fig:defect_residuals}       
\end{figure}

The first method that explicitly exploits the presence of strong in-field sensor defects to estimate the age of a digital image is proposed in \cite{Fridrich11a}. A defect is considered as being noise. Thus, by applying a denoising filter, the defect is filtered out. The median filter is a denoising filter that replaces the center value with the median of the neighboring pixel values. Typical kernel sizes are $3 \times 3$ or $5 \times 5$. Since the defect is filtered out (smoothed out) by applying a median filter, the median filter residuals, subtracting the median-filtered image from the original image, contain the defect magnitude (i.e., $IK + \tau D + c$). An example of median filter residuals of a defective pixel over chronologically ordered images is illustrated in Figure \ref{fig:defect_residuals}.

In \cite{Fridrich11a}, the assumption is that the difference between the noise residual and the sum of all defect parameters ($K,D,c$) is normally distributed with a mean of zero and a variance $\sigma^2$. A maximum likelihood approach is proposed to estimate the parameters $(K, D, c, \sigma)$ before and after the defect onset, as well as the defect onset time $j$ (the index of the chronologically ordered trusted set). The age of an image under investigation is again approximated by a maximum likelihood approach, i.e.
\begin{equation}
\scriptstyle
\hat{j} = \argmax\limits_{j} \prod_{i \in \Omega} \frac{1}{\sqrt{2 \pi}\hat{\sigma}^{(\Psi)}_{(i)}} \exp{\frac{W_{(i)}-\left( I_{(i)} \hat{K}^{(\Psi)}_{(i)} + \tau \hat{D}^{(\Psi)}_{(i)} + \hat{c}^{(\Psi)}_{(i)}\right) }{2\hat{\sigma}^{2(\Psi)}_{(i)}}},
\label{eq:ml_ap_fridrich}
\end{equation}
where $\Omega$ denotes the set of all defective pixels and $W_{(i)}$ is the residual value of the $i^{\text{th}}$ defective pixel. The estimated defect parameters are regarded either before or after the defect onset, i.e., $\Psi = 1$ if $j$ is greater as the estimated onset time for the $i^{\text{th}}$ defective pixel. In other words, the approximated acquisition time $\hat{j}$ represents the index at which the difference between the residual value $W_{(i)}$ and the sum of the defect parameters $( I_{(i)} K^{(\Psi)}_{(i)} + \tau D^{(\Psi)}_{(i)} + c^{(\Psi)})$ is minimum for all $i \in \Omega$.

The proposed approach is evaluated based on images from three different imagers. The images cover a time period of 802 days, 649 days, and 1411 days, with a reported Mean Absolute Error (MAE) of 76.97, 113.3, and 140.49, respectively. However, since the accuracy of the proposed method obviously depends on the number of new defect onsets, the authors also report a relative estimation error. The relative estimation error is the ratio between the MAE and the average time between the defect onsets. The corresponding relative estimation errors are 1.71, 0.87 and 0.60.

\subsection{A Machine Learning Approach to Approximate the Age of a Digital Image}
\label{sec:Joechl20a}
Instead of predicting a certain index of the set of chronologically ordered trusted images, we consider image age approximation as a multi-class classification problem in \cite{Joechl20a}. This is reasonable because the age approximation is based on the presence of in-field sensor defects, and thus the temporal prediction accuracy is bounded by the time intervals between two consecutive defect onsets. Hence, the classes are defined by the defect onset times and the available trusted images. To approximate (classify) the age of an image, different machine learning techniques are evaluated (i.e., a Naive Bayes Classifier (NB) and a Support Vector Machine (SVM)) in \cite{Joechl20a}.

As classification features, the median filter residuals of defective pixels are regarded. To apply a NB in this context, the conditional probability distributions of residuals given the defect is present or not have to be estimated for each defective pixel. This is done using three different approaches: (i) by assuming the distribution follows a normal distribution (NB-NE), (ii) by a histogram estimation (NB-HE), (iii) and by a Kernel Density estimation (NB-KDE). Images of three different imagers are used for evaluation and the proposed machine learning methods are compared with the maximum likelihood approach proposed in \cite{Fridrich11a}.

In conclusion, the SVM achieves the best classification accuracy and is the recommended approach. Comparing the different probability density estimation techniques, the best results are achieved by the NB-HE approach. However, this technique tends to overfit the data. In general, when a class is defined by multiple defects, it is more likely that not all defects are attenuated due to local scene properties. The classification performance obtained for such classes is very high (i.e., a f1 score of 0.9976 for Canon class 7 and 0.9227 for the final Nikon class).\\

Prior to applying techniques that exploit in-field sensor defects for image age approximation (i.e., \cite{Fridrich11a, Joechl20a}), the defect locations have to be determined. To this end, we introduce in \cite{Joechl21a} a method for defect detection, designed specifically in the context of image age approximation. General methods for detecting defective pixels are proposed, for example, in \cite{Chan09a,Cho11a,Chen12a,ElYamany17a,Tchendjou20a,Ghosh08a}. The effect of image compression on defect detection and in-field sensor defect based image age approximation is evaluated in \cite{Joechl21c}.

\subsection{A Machine Learning-based Approach for Picture Acquisition Timeslot Prediction Using Defective Pixels}
\label{sec:Ahmed21a}
A machine learning approach that combines defect detection and age approximation into one method is introduced in \cite{Ahmed21a}. With this technique, an arbitrary but fixed image block is first selected (e.g., a block of $200 \times 200$ pixels). Then, a classifier (e.g., K-Nearest Neighbor (KNN)) is trained for each pixel (except border pixel) inside the image block, i.e., each classifier corresponds to a specific pixel location. Given a $200 \times 200$ pixel block, this results in 39204 (i.e., $200 \times 200 - 796 \: (\text{border pixels}) = 39204$) trained classifiers, each predicting an age class. As features for training a specific classifier, the raw pixel values in a $w \times w$ (i.e., $3 \times 3$) window around the pixel under consideration are used (i.e., 27 pixel values, 9 per color channel). In addition, two dedicated local variation features are added per color channel to mimic the behavior of defective pixels. These features are defined as follows:
\begin{eqnarray}
 & LV1 = \text{abs}(x_c - avg), \label{eq:Ahmed21a01}\\
 & \text{where} \quad avg = \frac{\sum_{i=1}^{w}\sum_{j=1}^{w} x_{i,j}-x_c}{w \times w -1}, \nonumber \\
 & LV2 = \text{sqrt}(\frac{\sum_{i=1}^{w}\sum_{j=1}^{w} (x_{i,j}-x_c)^2}{w \times w -1}), \label{eq:Ahmed21a02}
\end{eqnarray}
where $x_c$ denotes the center pixel value. This results in a 33 dimensional feature vector, including 27 raw pixel values and 6 local variation features, used for training the classifier. The K (i.e., K = 100) classifiers with the best performance on the validation-set are selected. These classifiers, or rather the corresponding pixels, represent defective pixels. Then these 100 classifiers are retrained on the train- and validation-set. However, virtual sub-classes are used for retraining, with each class divided into two sub-classes. This poses a more challenging problem, which should boost the performance. In the evaluation stage, the prediction of the virtual sub-class is reconstructed into the actual class. A majority voting of all 100 classifiers determines the class prediction of the image block. The final class prediction is obtained by a majority vote on the predictions of multiple non-overlapping image blocks (i.e., 45 blocks).

\begin{table}
\begin{center}
\caption{Overview of the imagers used to create the NTIF dataset.}
\label{tab:NTIF_imagers}
\small
\begin{tabular}{ l | l | c }
ID & Model & Sensor \\
\hline \hline
Nc01 & Canon IXUS115HS & CMOS \\ \hline
Nc02 & Canon IXUS115HS & CMOS \\ \hline
Nf01 & Fujifilm S2950 & CCD \\ \hline
Nf02 & Fujifilm S2950 & CCD \\ \hline
Nn01 & Nikon Coolpix L330 & CCD\\ \hline
Nn02 & Nikon Coolpix L330 & CCD\\ \hline
Np01 & Panasonic DMC TZ20 & CMOS \\ \hline
Np02 & Panasonic DMC TZ20 & CMOS \\ \hline
Ns01 & Samsung pl120 & CCD \\ \hline
Ns02 & Samsung pl120 & CCD \\ \hline
\hline
\end{tabular}
\end{center}
\end{table}

The authors evaluate the proposed approach based on the publicly available Northumbria Temporal Image Forensics (NTIF) \cite{Ahmed20a} database. An overview of the imagers used to create the NTIF dataset is given in Table \ref{tab:NTIF_imagers}. Images from this database are also used in other image forensics related publications, i.e., \cite{Alani17a, Lawgaly17a}. For evaluation the first 40 weeks (time-slots) are divided into 5 classes (8 weeks per class). 8 weeks per class are chosen on the basis of a statement in \cite{Fridrich11a} that defective pixels occur in a period of two months. In total, three different classifiers are evaluated: a KNN classifier, a Naive Bayes classifier and an SVM. The best results are obtained with the KNN (up to 93\% accuracy for a 5 class problem).

\subsection{Temporal Image Forensic Analysis for Picture Dating with Deep Learning}
\label{sec:Ahmed20b}
In contrast to traditional machine learning techniques, a deep neural network learns the classification features used. For this reason, all existing age traces introduced by the image acquisition pipeline can be exploited by the network. Such rich age features (e.g., compared to features based only on the presence of in-field sensor defects) have the potential to improve the classification accuracy and temporal resolution significantly.

The first approach utilizing deep neural networks for image age approximation is proposed in \cite{Ahmed20b}. The authors utilize two well-known CNN architectures for this purpose, i.e., the AlexNet\cite{AlexNet} and GoogLeNet \cite{GoogleNet}. In particular, the two pre-trained models are fine-tuned according to the target task of image age approximation (classification). This is done based on two different modes: the transfer learning mode and the feature extraction mode. In transfer learning mode, the CNN is trained to output the target class directly. The classification head of the CNN is omitted in the feature extraction mode. In this mode, the features are fed into an additional machine learning classifier (i.e., a KNN and a SVM). Feature extraction mode is performed using AlexNet only.

To enlarge the number of training images, each image is cropped into 48 non-overlapping blocks of size $500 \times 500$. At test time, each test (query) image is also cropped into 48 non-overlapping blocks, and a majority vote of each image block prediction is performed to obtain the final class prediction. The proposed approach is also evaluated based on images from the NTIF database. However, this time only images of five imagers are considered and the first 25 time-slots are divided into 5 classes. In general, the transfer learning mode performs better than the feature extraction mode, with AlexNet outperforming GoogLeNet in the transfer learning mode on 4 out of 5 evaluated imagers (up to 88\% accuracy for a 5 class problem). Since the networks are trained on multiple non-overlapping image blocks, the authors suggest that the features learned are not dependent on block location.

\subsection{Temporal Image Forensics: Using CNNs for a Chronological Ordering of Line-Scan Data}
\label{sec:linescan_age_approximation}
CNN-based image dating of palmprints captured with a line scanner is performed in \cite{Paulitsch21a}. In principle, we proposed a multi-channel approach, where a modified version of the AlexNet \cite{AlexNet} is utilized per channel. All channels are combined by a classification head. The input for each channel are non-overlapping blocks extracted from a query image. For this purpose, different block extraction strategies are proposed. For example, blocks are extracted at fixed image locations (e.g., at the image corners and center, five crop). In addition to extraction at fixed positions, a so-called `Line mode' is proposed, in which the image is divided into consecutive, non-overlapping lines of the same height. Since a line scanner is used for image acquisition, the age signal present should be similar in each line. For each line $n$ blocks are extracted and fed into the $n$-channel model. The blocks are always extracted at the same relative position in each line, i.e., the $i$-th block representing the $i$-th channel should contain the same age signal pattern, independent of which line it is extracted from. The same strategy is also performed in a column based mode. Furthermore, a single-channel approach is proposed, where each block is sequentially forwarded to the same channel, i.e., position-dependent features can no longer be learned efficiently.

\begin{figure*}[!htb]
\centering
\includegraphics[width=0.88\textwidth]{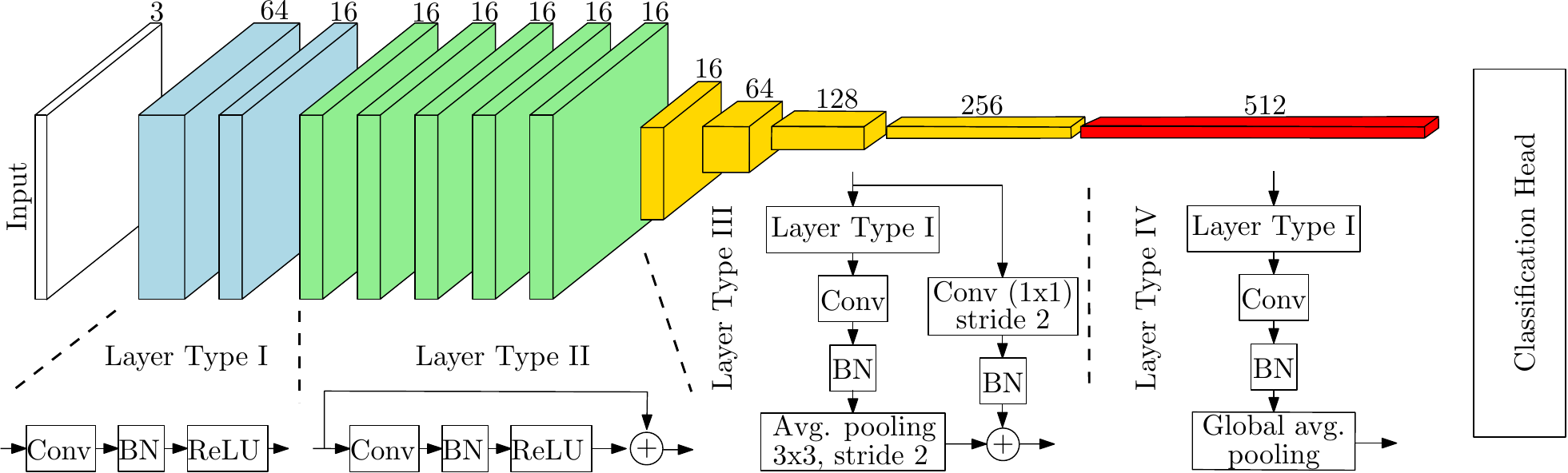}
\caption{Overview of the SRNet architecture, relying on the illustration in \cite{Boroumand18a}.}
\label{fig:SRNet}       
\end{figure*}

The approach is trained and evaluated using a binary classification problem with a time difference of about 5 years between the classes. The best performance (i.e., 100\% classification accuracy) is achieved with all fixed position block extraction strategies and a multi-channel model. The classification accuracy significantly decreases when a single-channel model is used. This implies that position dependent features are exploited. The median performance of the line and column-based block extraction strategies is only slightly better than prediction by chance. The poor performance of the line based extraction strategy is surprising, as the age pattern generated by a line scanner should theoretically be repeated per scan line. This, and the observation that line and column-based block extraction had similar performance, suggests that age traces introduced by the capturing device are unlikely to be exploited. In the case of the five-crop block extraction strategy, the most discriminating block is the one extracted from the top left corner (background only, black only) and the worst is the one extracted from the center (foreground only, palm region).

\section{Learned Features}
\label{sec:features}
In \cite{Fridrich11a,Joechl20a}, hand-crafted features (i.e., median filter residuals at defect positions) are used to approximate the age of a digital image. When utilizing data-driven algorithms such as deep neural networks, the selection of features is no longer essential. The model itself decides which features (which correlations in the training data) are used to best discriminate the training data. For this reason, and because deep neural networks usually consist of millions of parameters, these models turn into a `black box'.

To investigate the features learned by a CNN trained in the context of temporal image forensics, we systematically analyze in \cite{Joechl21b}: (i) how relevant the exact position of a strong in-field sensor defect is for training a CNN in the context of image age approximation, (ii) are additional `age' traces other than strong sensor defects exploited by the network and (iii) are the learned features positionally invariant. For this purpose, we train a deep neural network, the Steganalysis Residual Network (SRNet) \cite{Boroumand18a}, from scratch based on different learning scenarios in combination with different cropping methods. The SRNet is based on the famous residual learning principle and depicted in Figure \ref{fig:SRNet}. The key part of the architecture are the first 7 layers (i.e., Layer Type I and Layer Type II) without a pooling operation. Pooling acts like low-pass filtering. Thus, omitting it does not suppress the noise-like stego (or age) signal. The evaluation of the different learning scenarios in combination with different cropping methods shows that the presence of strong in-field sensor defects is irrelevant for the SRNet to achieve the reported classification accuracy. For this reason, it is suggested that other (unknown) `age' traces are exploited by the network. These learned `age' features are also not positionally invariant.

In \cite{Joechl22a}, we further investigated whether these learned `age' features are entirely device dependent. For this purpose, the SRNet is trained on images from a specific device. Afterwards, the trained model is applied to approximate the age of images from different devices. This evaluation is performed on 14 different imagers, where 10 of these 14 imagers are from the NTIF database. Based on the results obtained no overall trend is observable. For example, it can be observed that models trained on images of Nf02, Np01 and Ns01 are basically fully device independent (at least for the same device model pair), while models trained on images of Pc01, Pp01 and Pp02 are not device independent. Thus, the question if the learned `age' features are device (in)dependent could not be answered. In contrast, the reported results cast doubt on whether the network actually exploits solely age-related features to predict age class. Age traces can be interpreted as a weak signal that is hidden in a digital image. Nevertheless, this inherent age signal might not be the only difference between the age classes of regular scene images. Other non-age-related correlations in the training data (e.g., due to image content) can distract the network from learning age-related features.

\section{Content Bias}
\label{sec:content_bias}
Images belonging to the same age class are usually taken in close temporal proximity. Thus, it is very likely that the images of a particular age class share common non-age-related features in addition to the age traces. Such non-age-related features could be due to the following: (i) common scene properties (e.g., urban or nature scenes); (ii) common weather conditions (e.g., cloudy or blue sky); (iii) seasonal commonalities (e.g., light conditions and vegetation). Since the age traces are usually very weak, they can easily be obscured by other non-age related correlations in the data. A CNN can exploit such non-age related features to discriminate between the age classes. For example, if the first age class images are taken on a ski day in winter and the second age class images are taken on a hiking trip in the summer, the network will most certainly learn that the first class images are primarily white and the second class images are green. These non-age-related correlations in the training data, which distract the network from exploiting the age traces, can be referred to as `content bias'.

\begin{figure}[!t]
 \centering
 \begin{minipage}[b]{0.22\textwidth}
	\centering
	\includegraphics[width=0.94\textwidth]{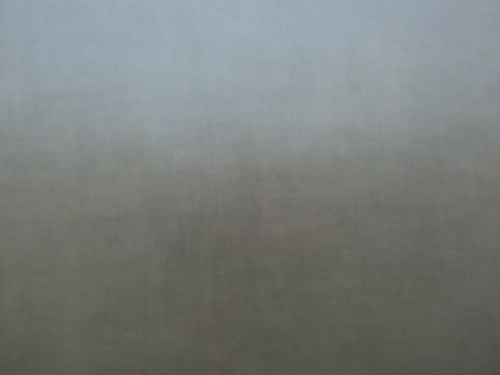}
	\centerline{\small{class 1}}\medskip
\end{minipage}
\begin{minipage}[b]{0.22\textwidth}
	\centering
	\includegraphics[width=0.94\textwidth]{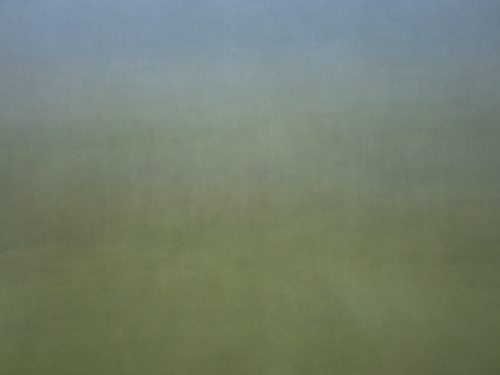}
	\centerline{\small{class 2}}\medskip
\end{minipage}
\caption{Example of average images, as illustrated in \cite{Joechl23a}, it is clearly visible that class 2 images probably contain more nature scenes (more green).}
\label{fig:example_content_bias_Joechl23a}
\end{figure}

An example of content bias is given in Figure \ref{fig:example_content_bias_Joechl23a}, where average images of all samples of a certain class are shown. In particular, the average images illustrated are generated from two classes of the NTIF dataset as used in \cite{Ahmed21a,Ahmed20b,Joechl22a,Joechl22b,Joechl23a}. The inherent content bias is clearly visible, i.e., the class 2 samples are likely to contain more nature scenes (more green). Apart from the recorded scenes, different image compression settings also leave certain traces in an image. For example, in \cite{Cattaneo14a} it is shown that a method for image forgery detection has actually learned to detect the different image quality factors with which pristine and forged images are stored. Furthermore, the camera parameter can also influence the image characteristics, e.g., a higher ISO or a longer exposure time leads to more noise, while variations in focal length can lead to different distortions. A bias in the camera parameter settings can also be a result of the captured scenes (e.g., lighting conditions). As described in \cite{Joechl23a}, a bias in the camera parameters can be observed for the NTIF classes.

\subsection{An Example of How easily a Neural Network can be distracted by Content Bias}
\label{sec:example}
\begin{figure*}[!t]
	\centering
	\begin{minipage}[b]{0.32\textwidth}
		\centering
		\includegraphics[width=0.88\linewidth]{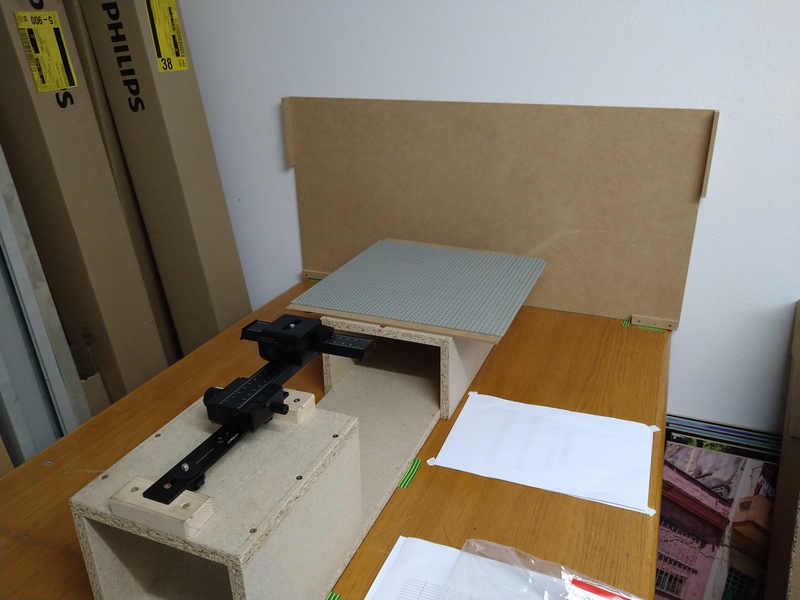}
		\centerline{(a) acquisition fixture}\medskip
	\end{minipage}
	\begin{minipage}[b]{0.32\textwidth}
		\centering
		\includegraphics[width=0.88\linewidth]{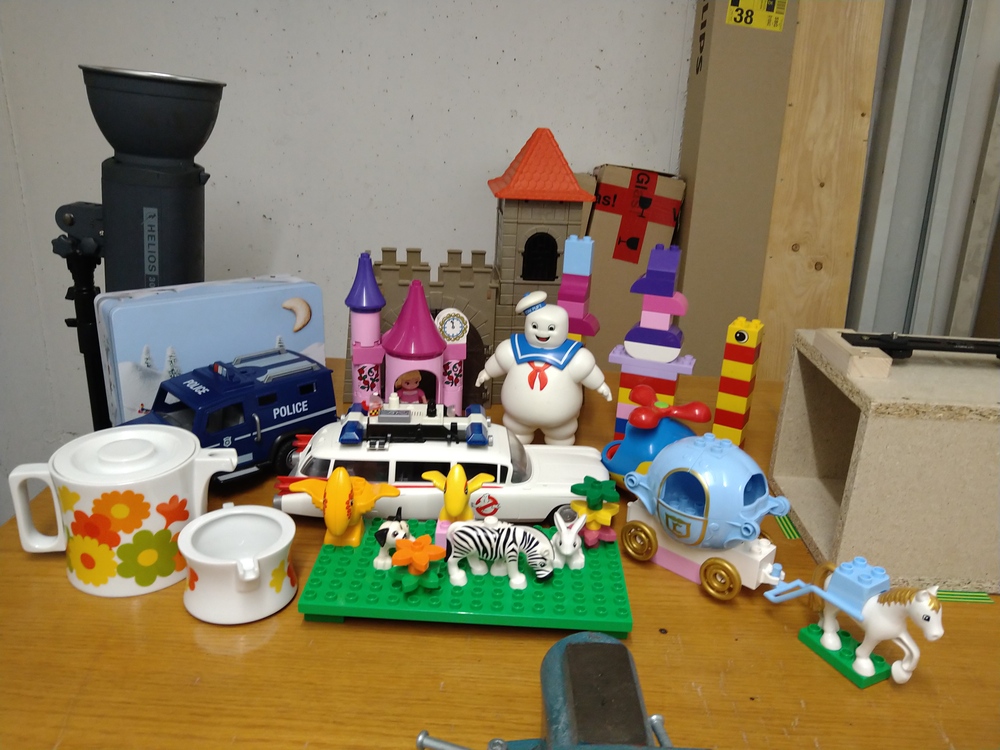}
		\centerline{(b) objects}\medskip
	\end{minipage}
	\begin{minipage}[b]{0.32\textwidth}
		\centering
		\includegraphics[width=0.98\linewidth]{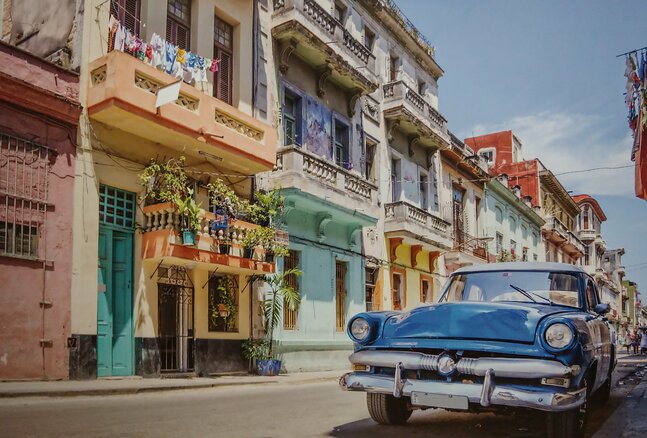}
		\centerline{(c) background example}\medskip
	\end{minipage}
	\caption{Overview of the used acquisition fixture (a), the available objects (b) and one background example (c).}
	\label{fig:std_scene_setup}
\end{figure*}

Given the problem of content bias when training a neural network for an age classification problem, the idea was to create a special dataset where content bias should be minimized. This could be achieved by capturing standardized scenes, as was done when creating the Dresden DB \cite{Gloe09a}. However, in the context of image age approximation, the same standardized scenes should be captured for each time-slot. The goal in capturing standardized scene images is to always capture the same image content at different points in time. This is to ensure that there are no inter-class differences other than potential existing age traces. For this purpose, standardized scene images (permutations of different backgrounds and foreground objects) are acquired in a controlled environment at different points in time. The used acquisition fixture, the foreground objects and background example are illustrated in Figure \ref{fig:std_scene_setup}. Interested readers will find the detailed protocol for acquiring these standardized scene images in the supplementary material.

Once images from two sessions (time difference of 4 months) and three devices were available, a SRNet was trained to discriminate the captured images from both sessions. In particular, five different SRNets are trained on five different fixed image blocks ($256 \times 256$), where each image block is always extracted from the same location (i.e., top left (tl), top right (tr), bottom left (bl), bottom right (br) corners and the center of the image (ce)). The final class prediction is obtained by fusing together the different scores of each individual sub-network. Furthermore, training is performed using two different sampling strategies: (i) random stratified sampling, (ii) background strict sampling. With the random stratified sampling strategy, data are drawn randomly for each age-class. Therefore, there is a high probability that backgrounds and objects that are present in the training set also appear in the test-set. To ensure that no background present in the training set appears in the test-set, the background strict sampling method is applied. In total, 8 runs are performed for each sampling strategy, where 80\% of images per session are used for training.

\begin{figure*}[!t]
    \centering
    \begin{minipage}[b]{0.42\textwidth}
     \centering
     \includegraphics[width=0.9\textwidth]{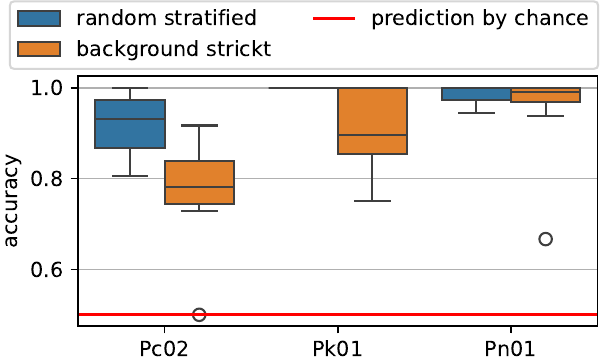}
     \centerline{\small{(a) general accuracy}}\medskip
    \end{minipage}
    \begin{minipage}[b]{0.57\textwidth}
     \centering
     \includegraphics[width=0.98\textwidth]{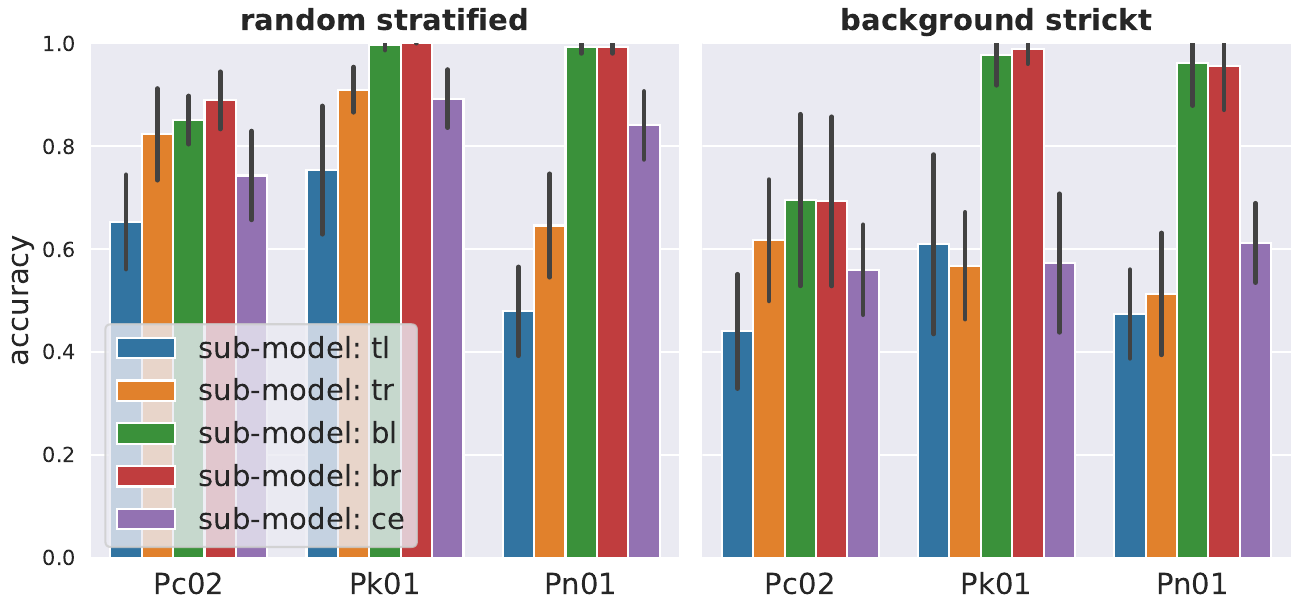}
     \centerline{\small{(b) accuracy per sub-model}}\medskip
    \end{minipage}
\caption{Classification accuracies of the acquired standard scene images. The classification accuracy achieved when the scores of all five sub-models are fused together are shown in (a), and (b) shows the accuracies achieved with the individual sub-models.}
\label{fig:eval_std_scenes}       
\end{figure*}

In Figure \ref{fig:eval_std_scenes} (a), boxplots of the classification accuracy values achieved for the random stratified and the background strict sampling methods are illustrated. It can be seen that with the random stratified strategy, the classification is constantly perfect with the Pk01, almost constantly perfect with the Pn01 and also very good with the Pc02. However, when background strict sampling is applied, the accuracy values decrease. In particular, for the Pc02, there is a run where the classification no longer is possible. If the network would have learned to detect solely age-related features, such a decrease in accuracy values between the different sampling strategies would not be expected. It could be that with some backgrounds, the detection of the age signal is more difficult than with others. Using the background strict strategy, such a `hard' background could then be included in the test-set. Nevertheless, this would rather explain single outliers than a general decrease.

For a more detailed analysis, the average accuracy values of the individual sub-models are illustrated in Figure \ref{fig:eval_std_scenes} (b). It can be seen that the two sub-models trained on image blocks extracted from both bottom corners (bl \& br) show the best performance on all devices. In case of the background strict sampling and for the Pn01 and Pk01, the performance of these two sub-models remains relatively the same as compared to the random stratified sampling strategy, while it decreases significantly for all other sub-models. One explanation would be that these two sub-models indeed learned age-related features. To confirm this assumption, the class activation maps of these two sub-models are analyzed.

CAMs are saliency maps that indicate the discriminative region used by a CNN to identify a certain class. GradCAM++ \cite{Chattopadhay18a}, based on the implementation in \cite{GildenblatPytorchCAM}, is used to analyze the trained sub-models.

\begin{figure*}[!t]
    \centering
    \begin{minipage}[b]{0.18\textwidth}
        \centering
        \includegraphics[width=0.88\textwidth]{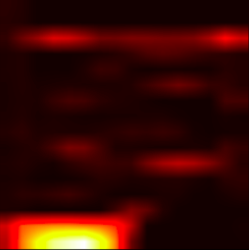}
    \end{minipage}
    \begin{minipage}[b]{0.18\textwidth}
        \centering
        \includegraphics[width=0.88\textwidth]{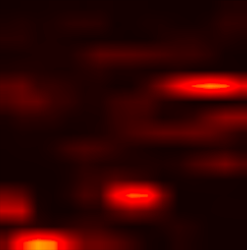}
    \end{minipage}
    \begin{minipage}[b]{0.18\textwidth}
        \centering
        \includegraphics[width=0.88\textwidth]{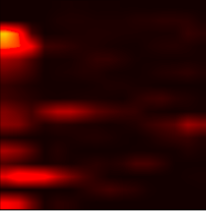}
    \end{minipage}
    \begin{minipage}[b]{0.18\textwidth}
        \centering
        \includegraphics[width=0.88\textwidth]{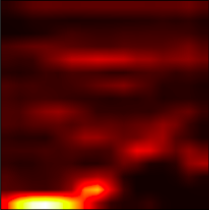}
    \end{minipage}
    \begin{minipage}[b]{0.18\textwidth}
        \centering
        \includegraphics[width=0.88\textwidth]{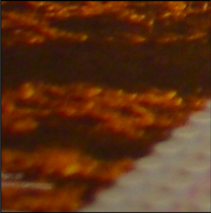}
    \end{minipage}
    \begin{minipage}[b]{0.18\textwidth}
        \centering
        \includegraphics[width=0.88\textwidth]{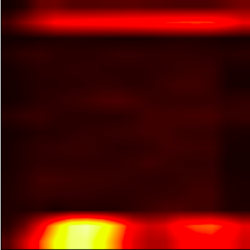}
        \centerline{\small{(a) run01}}\medskip
    \end{minipage}
    \begin{minipage}[b]{0.18\textwidth}
        \centering
        \includegraphics[width=0.88\textwidth]{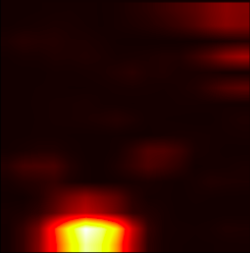}
        \centerline{\small{(b) run02}}\medskip
    \end{minipage}
    \begin{minipage}[b]{0.18\textwidth}
        \centering
        \includegraphics[width=0.88\textwidth]{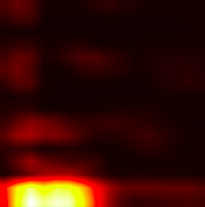}
        \centerline{\small{(c) run03}}\medskip
    \end{minipage}
    \begin{minipage}[b]{0.18\textwidth}
        \centering
        \includegraphics[width=0.88\textwidth]{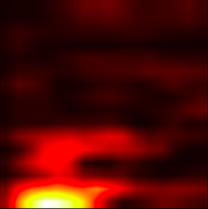}
        \centerline{\small{(d) run04}}\medskip
    \end{minipage}
    \begin{minipage}[b]{0.18\textwidth}
        \centering
        \includegraphics[width=0.88\textwidth]{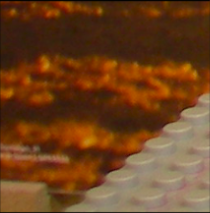}
        \centerline{\small{(e) example image block}}\medskip
    \end{minipage}
    \caption{Superimposed CAMs for correctly predicted test samples. The model is trained with image blocks extracted from the bottom left corner of Pn01 images (top-row: session 1, bottom-row: session 2).}
    \label{fig:cam_overlay_std_scene_f-id02}
\end{figure*}
\begin{figure*}[!t]
    \centering
    \begin{minipage}[b]{0.18\textwidth}
        \centering
        \includegraphics[width=0.84\textwidth]{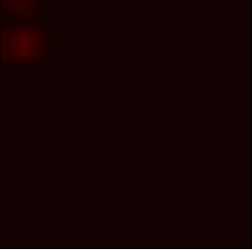}
    \end{minipage}
    \begin{minipage}[b]{0.18\textwidth}
        \centering
        \includegraphics[width=0.84\textwidth]{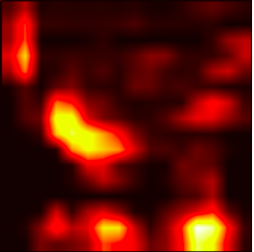}
    \end{minipage}
    \begin{minipage}[b]{0.18\textwidth}
        \centering
        \includegraphics[width=0.84\textwidth]{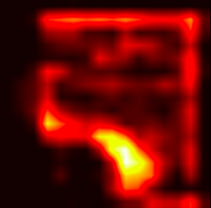}
    \end{minipage}
    \begin{minipage}[b]{0.18\textwidth}
        \centering
        \includegraphics[width=0.84\textwidth]{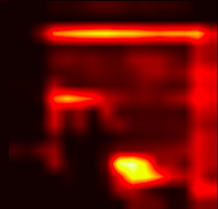}
    \end{minipage}
    \begin{minipage}[b]{0.18\textwidth}
        \centering
        \includegraphics[width=0.84\textwidth]{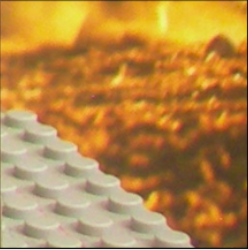}
    \end{minipage}
    \begin{minipage}[b]{0.18\textwidth}
        \centering
        \includegraphics[width=0.84\textwidth]{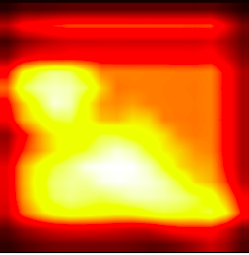}
        \centerline{\small{(a) run01}}\medskip
    \end{minipage}
    \begin{minipage}[b]{0.18\textwidth}
        \centering
        \includegraphics[width=0.84\textwidth]{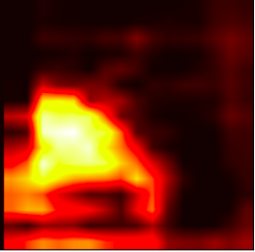}
        \centerline{\small{(b) run02}}\medskip
    \end{minipage}
    \begin{minipage}[b]{0.18\textwidth}
        \centering
        \includegraphics[width=0.84\textwidth]{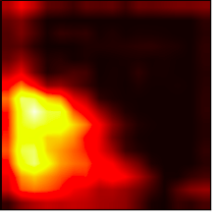}
        \centerline{\small{(c) run03}}\medskip
    \end{minipage}
    \begin{minipage}[b]{0.18\textwidth}
        \centering
        \includegraphics[width=0.84\textwidth]{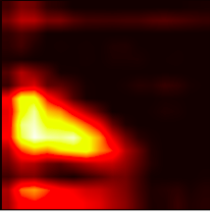}
        \centerline{\small{(d) run04}}\medskip
    \end{minipage}
    \begin{minipage}[b]{0.18\textwidth}
        \centering
        \includegraphics[width=0.84\textwidth]{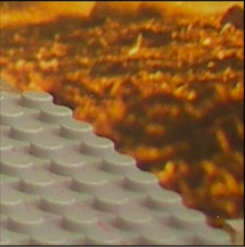}
        \centerline{\small{(e) example image block}}\medskip
    \end{minipage}
    \caption{Superimposed CAMs for correctly predicted test samples. The model is trained with image blocks extracted from the bottom right corner of Pn01 images (top-row: session 1, bottom-row: session 2).}
    \label{fig:cam_overlay_std_scene_f-id03}
\end{figure*}
In Figure \ref{fig:cam_overlay_std_scene_f-id02} and \ref{fig:cam_overlay_std_scene_f-id03} the superimposed class activation maps of correctly predicted test samples, of the first 4 runs from the Pn01 and the sub-models trained on images blocks extracted from the bottom left and right corner of the image, are illustrated. The top rows in both Figures (\ref{fig:cam_overlay_std_scene_f-id02} and \ref{fig:cam_overlay_std_scene_f-id03}) show the activation maps of session 1 (class 1) and the bottom row of session 2 (class 2). In Figure \ref{fig:cam_overlay_std_scene_f-id02} it is clearly visible that the main activation is on a small piece of wood at the bottom left corner of the image block. This piece of wood is used as background fixation and is clearly visible in images from session 2. For the model trained on image blocks extracted from the bottom right corner of the image, a similar phenomenon can be observed. In this case, a larger area of the object plate can be seen in session 2, which is also reflected by the superimposed class activation maps. Similar behavior can also be observed with the Pk01.

The observed differences between session 1 and 2 can be explained by a different vertical tilt of the camera. This tilt is caused by the fact that the camera release plate is not directly pillowed from below, which led to this tilt over time. Hence, the classification performance (of both sub-models trained on both bottom image blocks bl and br), which is independent of the sampling strategy, cannot be explained by a learned age signal but by the camera tilt. Because of a different resolution and a different lens with the Pc02, only background is visible in both bottom image blocks. This explains the constant lower performance of all 5 sub-models with the background strict sampling strategy.

The tilt of the camera is now fixed by adding a pillar. However, this example shows how easy a neural network can be distracted from learning an age signal. Particularly, as the age signal is very weak, it can easily be obscured by other (non-age-related) correlations in the training data. In this case, the introduced bias was obvious and easy to identify, but this may not always be the case. \\

The problem of deep neural networks learning an unintended bias from the training data is not unique in the context of temporal image forensics, but is a general problem. For example, a deep neural network was successfully trained in \cite{Khan20a} to detect a COVID-19 infection based on chest x-ray images. In \cite{Dhar21a}, however, it is shown that with the same X-ray images a successful classification (well above prediction by chance) is also possible on the basis of images blocks (upper left corner) showing only background. This does not directly imply that the decision of the model in \cite{Khan20a} is not based on reasonable features. However, if a successful classification (better than random prediction) is possible only based on the background, the model in \cite{Khan20a} might be biased and no valid conclusions can be drawn. This example is also highlighted in a Nature article, which poses the question: ``Is AI leading to a reproducibility crisis in science?''\cite{Ball23a}. An example from the field of image forensics is given in \cite{Mostafa25a}. It is shown that a method introduced for camera identification most likely learns content-dependent or color-dependent features to discriminate between devices.

These examples show that due to the `black box' nature of deep neural networks and their decisions, which are difficult to explain and interpret, incorrect or misleading conclusions can be drawn. Thus, methods from the field of XAI are of paramount importance for assessing the reliability of deep learning-based age approximation methods.

\section{XAI and Temporal Image Forensics}
\label{sec:explain}
The field of XAI focuses on understanding and interpreting the decisions made by deep neural networks. Comprehensive surveys from the field of XAI and trustworthiness AI is given in \cite{Linardatos21a,Arrieta20a,Ivanovs21a,Nima23a,Kemmerzell25a}. In the context of deep learning age approximation, the main objective (before focusing on understanding and interpreting the learned age features) is to assess whether the decision made is based on comprehensible evidence (i.e., a hidden age signal) or on an exploited content bias.

\begin{figure*}[!t]
	\centering
	\begin{minipage}[b]{0.32\textwidth}
		\centering
		\includegraphics[width=1\linewidth]{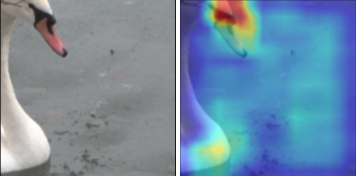}
		\centerline{(a)}\medskip
	\end{minipage}
	\begin{minipage}[b]{0.32\textwidth}
		\centering
		\includegraphics[width=1\linewidth]{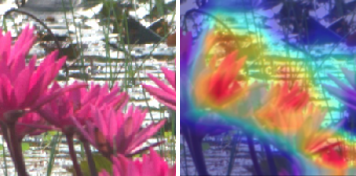}
		\centerline{(b)}\medskip
	\end{minipage}
	\begin{minipage}[b]{0.32\textwidth}
		\centering
		\includegraphics[width=1\linewidth]{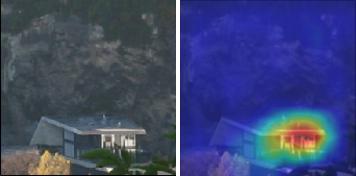}
		\centerline{(c)}\medskip
	\end{minipage}
	\begin{minipage}[b]{0.32\textwidth}
		\centering
		\includegraphics[width=1\linewidth]{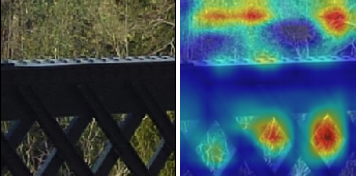}
		\centerline{(d)}\medskip
	\end{minipage}
	\begin{minipage}[b]{0.32\textwidth}
		\centering
		\includegraphics[width=1\linewidth]{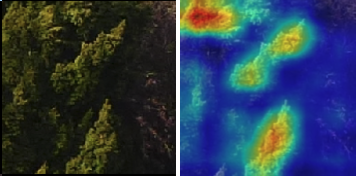}
		\centerline{(e)}\medskip
	\end{minipage}
	\begin{minipage}[b]{0.32\textwidth}
		\centering
		\includegraphics[width=1\linewidth]{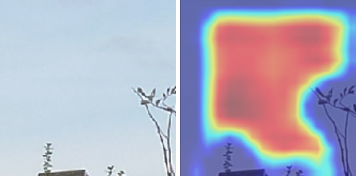}
		\centerline{(f)}\medskip
	\end{minipage}
	\caption{Example of activations directly on objects, on shrub-, tree-like structures and on areas, as illustrated in \cite{Joechl22b}. The color indicates the importance of a region, red being very important.}
	\label{fig:act_objects}
\end{figure*}

We analyse the features learned from a standard CNN trained on a binary age classification problem in \cite{Joechl22b}. For this purpose, methods from the field of XAI are applied to investigate whether age related features are learned. In particular, CAMs (i.e., the GradCAM++ \cite{Chattopadhay18a} and ScoreCAM \cite{WangH20a}) are used to analyse the features learned in \cite{Joechl22b}. As shown in Figure \ref{fig:act_objects}, activations directly on objects (Figure \ref{fig:act_objects} (a)-(c)), on shrub-like or tree-like structures (Figure \ref{fig:act_objects} (d) \& (e)) and on large homogeneous areas such as sky regions (Figure \ref{fig:act_objects} (f)) can be observed. If the model learned to exploit age-related features, activations independent of the image content (e.g., objects or sky regions) would be expected.

\begin{figure}[!t]
	\centering
	\begin{minipage}[b]{0.22\textwidth}
		\centering
		\includegraphics[width=0.92\linewidth]{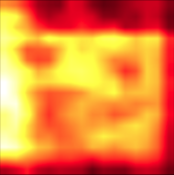}
		\centerline{run 1}\medskip
	\end{minipage}
	\begin{minipage}[b]{0.22\textwidth}
		\centering
		\includegraphics[width=0.92\linewidth]{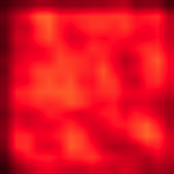}
		\centerline{run 2}\medskip
	\end{minipage}
	\begin{minipage}[b]{0.22\textwidth}
		\centering
		\includegraphics[width=0.92\linewidth]{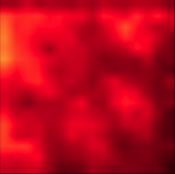}
		\centerline{run 3}\medskip
	\end{minipage}
	\begin{minipage}[b]{0.22\textwidth}
		\centering
		\includegraphics[width=0.92\linewidth]{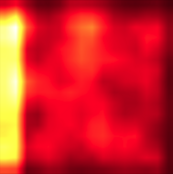}
		\centerline{run 4}\medskip
	\end{minipage}
	\caption{Examples of superimposed activations of correctly predicted image blocks of a given run, as illustrated in \cite{Joechl22b}.}
	\label{fig:act_overlay}
\end{figure}

Furthermore, a fairly constant activation pattern is expected across different input images when position-dependent age features, such as in-field sensor defects, are learned. To assess this, the superimposed activations of the correctly predicted test samples of a given test-set (run) are visualized. Since the image content varies between different sampled test-sets (runs), while the age signal should be constant, a similar activation pattern should be visible in the generated superimposed activations for all runs. An example is given in Figure \ref{fig:act_overlay}, where the superimposed activations of the first four runs are visualized. As can be seen, there is no constant activation pattern visible. This is only one example, but similar results are observed in all other models investigated. Based on these observations, the image content is probably more important for classification than the embedded age signal.

However, the analysis of CAMs is tedious and impractical when multiple models need to be evaluated. For this purpose, in \cite{Joechl23a} we present an approach in the context of XAI and temporal image forensics that evaluates the influence of image content.

\begin{figure}[!t]
	\centering
	\includegraphics[width=0.42\textwidth]{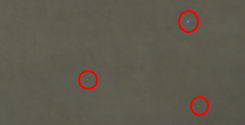}
    \caption{Example of average images, as illustrated in \cite{Joechl23a}, the inherent age signal (e.g., in-field sensor defects) is preserved.}
    \label{fig:examples_avg_Is_Joechl23a}
\end{figure}

To assess whether the model has learned to exploit solely age trace or image content, we propose to utilize average images in \cite{Joechl23a}. The proposed approach is based on the assumption that the inherent age signal is constant (except for small variations due to different camera settings) across all samples of a given age class. For this reason, when computing an average image across all samples of a given age class, the age signal should be preserved while image content is suppressed (as shown in Figure \ref{fig:examples_avg_Is_Joechl23a}). Thus, the classification accuracy of average images should be similar to the accuracy of the original input (as used for training) when inference is based (solely) on a hidden age signal. However, average images also represent the average image content.

\begin{figure}[!t]
    \centering
    \begin{minipage}[b]{0.104\textwidth}
        \centering
        \includegraphics[width=0.98\textwidth]{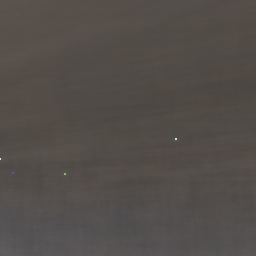}
        \centerline{\small{$\overline{Y^{k}}_{}$}}\medskip
    \end{minipage}
    \begin{minipage}[b]{0.104\textwidth}
        \centering
        \includegraphics[width=0.98\textwidth]{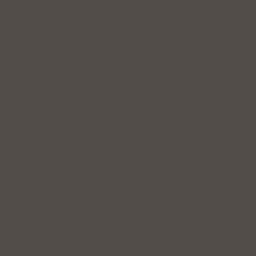}
        \centerline{\small{$\overline{Y^{k}}_{c}$ }}\medskip
    \end{minipage}
    \begin{minipage}[b]{0.104\textwidth}
        \centering
        \includegraphics[width=0.98\textwidth]{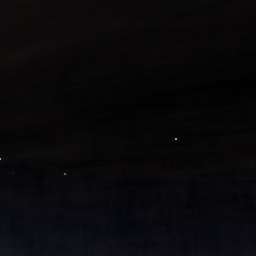}
        \centerline{\small{$\overline{Y^{k}}_{r}$}}\medskip
    \end{minipage}
    \begin{minipage}[b]{0.104\textwidth}
        \centering
        \includegraphics[width=0.98\textwidth]{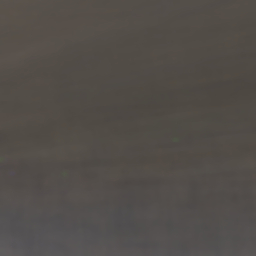}
        \centerline{\small{$\overline{Y^{k}}_{f}$}}\medskip
    \end{minipage}
    \caption{Examples of the different variants generated, as illustrated in \cite{Joechl23a}. In-field sensor defects are still present in $\overline{Y^{k}}$ and $\overline{Y^{k}}_{r}$.}
    \label{fig:avg_I_variants_Joechl23a}
\end{figure}

In total, it is proposed to create four different variants of average images per age class $k$:
 \begin{itemize}
  \item A standard average image, $\overline{Y^{k}}$.
  \item The average color of the average image (constant component), $\overline{Y^{k}}_{c}$. Since every pixel in $\overline{Y^{k}}_{c}$ has the same value, there is no age signal present anymore.
  \item The structure components, $\overline{Y^{k}}_{r}$, should only represent the age signal independent of the color.
  \item All high-frequency image components (e.g., in-field sensor defects) are filtered out in the median-filtered average image, $\overline{Y^{k}}_{f}$.
 \end{itemize}
 Examples of the different variants generated are illustrated in Figure \ref{fig:avg_I_variants_Joechl23a}.

The proposed XAI technique is a post-hoc method that aims to explain the behavior of a trained model at the time of testing. In other words, a model is trained on regular image data. The influence of image content on the discrimination performance is then evaluated by comparing the classification accuracy achieved when the original inputs (as used for training) and the different average images are provided.

Based on this XAI method, it is shown that the deep learning based image age approximation approach proposed in \cite{Ahmed20b} (see subsection \ref{sec:Ahmed20b}) most likely exploits content bias. Similarly, when the SRNet is trained in the context of image age approximation (as in \cite{Joechl21b,Joechl22a}, see section \ref{sec:features}), the inference is highly dependent on the image content. Even if pre-processing is applied to increase the signal-to-noise ratio (e.g., train on median filter residuals), the image content still plays an important role.

\section{XAI Evaluation of Tempor Image Forensics Methods}
\label{sec:xai_eval}
The observations in \cite{Joechl22b} and \cite{Joechl23a} stress and highlight the problem of content bias in the context of deep learning based image age approximation. However, content bias is not necessarily only a problem in deep learning based methods. The machine learning approach proposed in \cite{Ahmed21a} (see subsection \ref{sec:Ahmed21a}) combines defect detection and image age approximation in one method. Based on the design of the method, however, it is not evident that in-field sensor defects are actually exploited to obtain the reported classification accuracy values.

\subsection{XAI Evaluation of: A Machine Learning-based Approach for Picture Acquisition Timeslot Prediction Using Defective Pixels}
\label{sec:xai_Ahmed}
\begin{figure}[!t]
    \centering
    \includegraphics[width=0.48\textwidth]{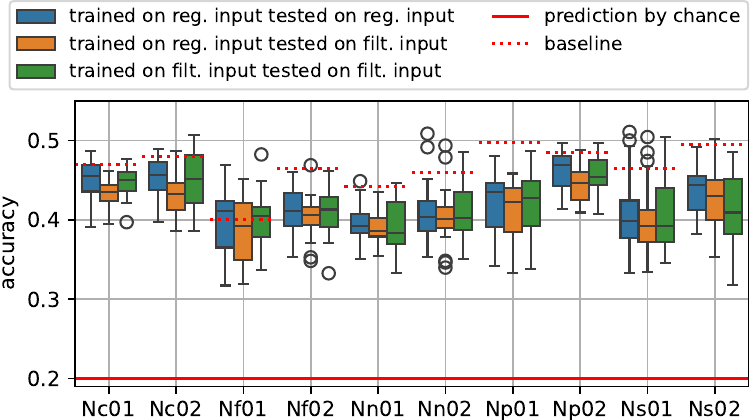}
  \caption{Boxplot of the 20 different accuracy values achieved per imager when the method was trained on 20 different image block selections along with 10 different train-, validation-, and test-sets.}
  \label{fig:results_Ahmed21a}
\end{figure}
Recall: In \cite{Ahmed21a} a classifier (i.e. KNN) is trained for each pixel (except the edge pixels) in a $200 \times 200$ image block. The 100 best classifiers or rather the corresponding pixels are regarded as defects. In other words, the 100 pixels with the most discriminative features (i.e., the raw pixel values around the pixel in question (27 values) and two local variation features (6 values), see equation (\ref{eq:Ahmed21a01}) and (\ref{eq:Ahmed21a02})) are considered as defects. A total of 45 non-overlapping image blocks are taken into account for the final class prediction.

For 100 classifiers per image block and 45 regarded image blocks, this results in 4500 classifiers or 4500 defective pixels. Since age approximation is supposed to be based on defective pixels, these 4500 defects have to occur within the considered time period (40 weeks). Such a defect growth rate is definitely not in line with the current literature (see section \ref{sec:traces}). Even if a very high number of defects develop during the regarded period, it is not guaranteed that all defects can be exploited by randomly selecting the locations of the used image blocks. Furthermore, the state of a defective pixel is basically binary (i.e., good or defect). In the proposed approach, a single classifier performs a five-class prediction (or 10 classes when using virtual sub-classes) based on a single defect. It is not evident from the literature that the defect parameters change over time in a way that would allow such a classification.

To additionally demonstrate that probably not in-field sensor defects are exploited, the first parts of the proposed method (i.e., without the virtual sub-classes and the selection of more than one image block) are implemented. It is reasonable to only implement the first parts, because the relevant part of feature generation is involved. The implemented method is then applied to the same data (NTIF) as in \cite{Ahmed21a}.

A boxplot of accuracy values obtained by 20 different image block selections along with 10 different train-, validation- and test-sets, is illustrated in Figure \ref{fig:results_Ahmed21a}. The baseline in Figure \ref{fig:results_Ahmed21a} represents the results reported in \cite{Ahmed21a}. In \cite{Ahmed21a}, it is not clearly defined how the reported accuracy values are obtained (e.g., average accuracy of different image block selection and/or different data sampling, or the maximum achieved accuracy, etc.). Nevertheless, the reported accuracy values can be verified based on the implemented method.

A median filter filters out (smoothes out) in-field sensor defects. For this reason, if the classifiers have learned to exploit in-field sensor defects, accuracy should decrease significantly when median-filtered images are presented to the trained classifiers. However, this can not be observed in Figure \ref{fig:results_Ahmed21a}. Furthermore, a similar accuracy can also be obtained when the method is trained on median-filtered images. This also shows that successfully learning an age (class) discrimination using this method is not dependent on the presence of in-field sensor defects.

To evaluate whether image content instead of defects was learned in \cite{Ahmed21a}, the XAI method described in section \ref{sec:explain} is applied to the trained classifiers. For this purpose, 20 different sets of average images are generated per class, with 80\% of all samples (including training, validation and testing data) randomly selected for each set. Table \ref{tab:xai_eval_Ahmed21a} shows the accuracies obtained for the original input $S$ (as used in training) and the accuracies obtained for the different average images.

\begin{table}[htb]
\centering
\caption{Accuracy values for the different input types. The first part of the machine learning approach proposed in \cite{Ahmed21a} is implemented and trained on the NTIF dataset (5 classes). Recall: An age signal is present in $\overline{Y}$ and $\overline{Y}_{r}$ and not in $\overline{Y}_{c}$. $\overline{Y}_{f}$, high frequency components (defects) are filtered out.}
\label{tab:xai_eval_Ahmed21a}
\begin{tabular}{l | c | c | c | c | c}
IDs & $\overline{Y}$ & $\overline{Y}_{c}$ & $\overline{Y}_{r}$ & $\overline{Y}_{f}$& $S$ \\ \hline \hline
Nc01 & 0.28 & 0.26 & 0.19 & 0.27 & 0.45 \\
Nc02 & 0.19 & 0.19 & 0.21 & 0.19 & 0.45 \\
Nf01 & 0.24 & 0.23 & 0.20 & 0.24 & 0.40 \\
Nf02 & 0.30 & 0.30 & 0.19 & 0.31 & 0.41 \\
Nn01 & 0.27 & 0.27 & 0.24 & 0.27 & 0.40 \\
Nn02 & 0.31 & 0.29 & 0.25 & 0.30 & 0.41 \\
Np01 & 0.23 & 0.21 & 0.21 & 0.23 & 0.42 \\
Np02 & 0.21 & 0.22 & 0.22 & 0.21 & 0.46 \\
Ns01 & 0.21 & 0.23 & 0.22 & 0.22 & 0.41 \\
Ns02 & 0.29 & 0.28 & 0.21 & 0.29 & 0.44 \\
\hline \hline
$\varnothing$ & 0.25 & 0.25 & 0.21 & 0.25 & 0.43 \\
\end{tabular}
\end{table}

If the model has learned to exploit an age signal, similar accuracy should be achieved if $\overline{Y}$ and $\overline{Y}_r$ or the original input $S$ (as used in training) is provided, since the age signal is retained in $\overline{Y}$ and $\overline{Y}_r$. A classification based on $\overline{Y}_c$ should not be possible, as $\overline{Y}_{c}$ is strongly dependent on image content. Furthermore, if features from the high-frequency image components are exploited (e.g., in-field sensor defects), classification based on $\overline{Y}_{f}$ should also not be possible.

As can be seen in Table \ref{tab:xai_eval_Ahmed21a}, the accuracy achieved based on $\overline{Y}$ is most of the time close to the random prediction border (i.e., average accuracy (across all devices) is $0.25 \approx 0.2$). This indicates that image content is probably exploited by the classifiers. Furthermore, if a correct classification of an average image ($\overline{Y}$) is possible, classification also works with the average color ($\overline{Y}_{c}$) and the median filtered average image ($\overline{Y}_{f}$). Since $\overline{Y}_{c}$ depends strongly on the image content, this is another indication that image content is exploited. Furthermore, a successful classification of $\overline{Y}_{f}$ additionally shows that in-field sensor defects are not utilized.

Based on the theoretical doubts and the empirical observations, it is not very likely that the method proposed in \cite{Ahmed21a} exploits the presence of in-field sensor defects to approximate the image age. Instead, content bias is probably being exploited.

\subsection{XAI Evaluation of: Temporal Image Forensics: Using CNNs for a Chronological Ordering of Line-Scan Data}
As described in subsection \ref{sec:linescan_age_approximation} we trained a CNN to discriminate images of palmprints captured with a line scanner. Since a line scanner is used, an age pattern (i.e., induced by the capture device) should be repeated per scan-line. However, this could not be observed in \cite{Paulitsch21a}. Only when the model is trained on blocks extracted from fixed image positions a successful discrimination was possible (i.e., 100\% accuracy for a binary classification problem). To further analyse the features learned, we re-implemented the proposed approach. Using the five-crop block extraction strategy combined with a multi-channel model, the results in \cite{Paulitsch21a} can be confirmed (except for 1/10 evaluated runs, the classification accuracy is not 100\%). The accuracy values that result when the different types of average images are applied to the trained models are listed in Table \ref{tab:xai_eval_Paulitsch21a}. It can be seen that the classification accuracy achieved on the basis of average images $\overline{Y}$ is identical to the original input $S$. Classification based on the average color ($\overline{Y}_c$) does not work. The median filtering of the average images ($\overline{Y}_f$) and the shifting of the pixel values ($\overline{Y}_r$) have a slight effect on the classification performance.

\begin{table}[htb]
\centering
\caption{Accuracy values for the different input types, multi-channel model combined with the five-crop block extraction strategy (2 classes).}
\label{tab:xai_eval_Paulitsch21a}
\begin{tabular}{ c | c | c | c | c} $\overline{Y}$ & $\overline{Y}_{c}$ & $\overline{Y}_{r}$ & $\overline{Y}_{f}$& $S$ \\ \hline \hline
0.95 & 0.50 & 0.90 & 0.85 & 0.95
\end{tabular}
\end{table}

\begin{figure}[!t]
    \centering
    \begin{minipage}[b]{0.088\textwidth}
        \centering
        \includegraphics[width=0.98\textwidth]{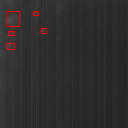}
        \centerline{\small{C0 - TL}}\medskip
    \end{minipage}
    \begin{minipage}[b]{0.088\textwidth}
        \centering
        \includegraphics[width=0.98\textwidth]{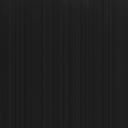}
        \centerline{\small{C0 - TR}}\medskip
    \end{minipage}
    \begin{minipage}[b]{0.088\textwidth}
        \centering
        \includegraphics[width=0.98\textwidth]{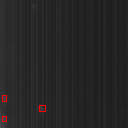}
        \centerline{\small{C0 - BL}}\medskip
    \end{minipage}
    \begin{minipage}[b]{0.088\textwidth}
        \centering
        \includegraphics[width=0.98\textwidth]{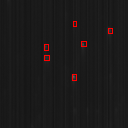}
        \centerline{\small{C0 - BR}}\medskip
    \end{minipage}
    \begin{minipage}[b]{0.088\textwidth}
        \centering
        \includegraphics[width=0.98\textwidth]{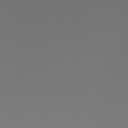}
        \centerline{\small{C0 - CE}}\medskip
    \end{minipage}

    \begin{minipage}[b]{0.088\textwidth}
        \centering
        \includegraphics[width=0.98\textwidth]{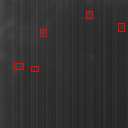}
        \centerline{\small{C1 - TL}}\medskip
    \end{minipage}
    \begin{minipage}[b]{0.088\textwidth}
        \centering
        \includegraphics[width=0.98\textwidth]{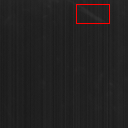}
        \centerline{\small{C1 - TR}}\medskip
    \end{minipage}
    \begin{minipage}[b]{0.088\textwidth}
        \centering
        \includegraphics[width=0.98\textwidth]{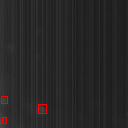}
        \centerline{\small{C1 - BL}}\medskip
    \end{minipage}
    \begin{minipage}[b]{0.088\textwidth}
        \centering
        \includegraphics[width=0.98\textwidth]{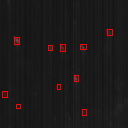}
        \centerline{\small{C1 - BR}}\medskip
    \end{minipage}
    \begin{minipage}[b]{0.088\textwidth}
        \centering
        \includegraphics[width=0.98\textwidth]{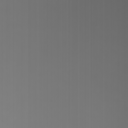}
        \centerline{\small{C1 - CE}}\medskip
    \end{minipage}

    \begin{minipage}[b]{0.088\textwidth}
        \centering
        \includegraphics[width=0.98\textwidth]{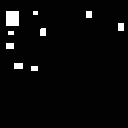}
        \centerline{\small{mask - TL}}\medskip
    \end{minipage}
    \begin{minipage}[b]{0.088\textwidth}
        \centering
        \includegraphics[width=0.98\textwidth]{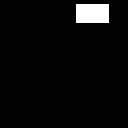}
        \centerline{\small{mask - TR}}\medskip
    \end{minipage}
    \begin{minipage}[b]{0.088\textwidth}
        \centering
        \includegraphics[width=0.98\textwidth]{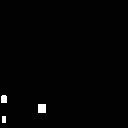}
        \centerline{\small{mask - BL}}\medskip
    \end{minipage}
    \begin{minipage}[b]{0.088\textwidth}
        \centering
        \includegraphics[width=0.98\textwidth]{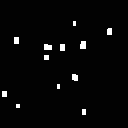}
        \centerline{\small{mask - BR}}\medskip
    \end{minipage}
    \begin{minipage}[b]{0.088\textwidth}
        \centering
        \includegraphics[width=0.98\textwidth]{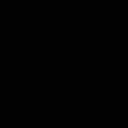}
        \centerline{\small{mask - CE}}\medskip
    \end{minipage}
    \caption{Example of the extracted blocks in the five-crop approach. The extraction locations are the top left (TL), top right (TR), bottom left (BL), bottom right (BR) corner and the center (CE). The first row represents the blocks extracted from an average image of class 0 (C0) and the second row from an average image of class 1 (C1). All red rectangles mark the position of dust/dirt spots. The last row represents the masks generated based on the marked positions.}
    \label{fig:Paulitsch21a_masks}
\end{figure}

Based on these observations, it could be assumed that an age signal is learned. However, the nature of the used line-scanner images is very different from real scene images (i.e., all four corner blocks contain only background (black) and the center block parts of the palm). All five extracted blocks are visualized in Figure \ref{fig:Paulitsch21a_masks}. The blocks are extracted from an average image generated for each class (C0 and C1). A visual inspection revealed a different pattern of small white spots on each block (highlighted by the red rectangles in Figure \ref{fig:Paulitsch21a_masks}). These small white spots are likely due to dust or dirt on the scanner's protective glass, and the resulting pattern is probably being exploited by the network.

\begin{figure}[!t]
	\centering
	\includegraphics[width=0.44\textwidth]{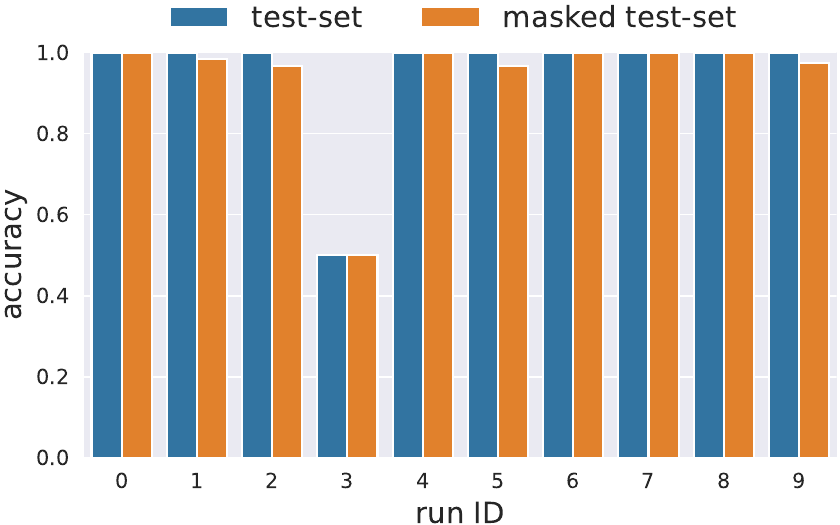}
    \caption{Accuracy achieved for each run based on a not-masked and masked test-set.}
    \label{fig:Paulitsch21a_xai_barplot}
\end{figure}

To verify this, a model trained with a regular (non-masked) input is evaluated with blocks where only the dust/dirt spots are retained. For this purpose, a mask is created in which solely the pixels in areas containing a dust/dirt spot are set to 1, all others to 0 (as shown in Figure \ref{fig:Paulitsch21a_masks}, last row). The mask is multiplied by the input block. This means that only dust/dirt spot areas are available as information that can be utilized by the trained network for classification. The resulting accuracy values for each run are illustrated in Figure \ref{fig:Paulitsch21a_xai_barplot}. As can be seen, the accuracy achieved with the masked input is just as or almost as high as with the unmasked input. Based on this result, it can be concluded that the classification is most likely largely based on the dust/dirt pattern. This dust/dirt pattern can be regarded as an age signal (similar to sensor dust, as described in section \ref{sec:sensor_dust}). However, since only the background (black area) is shown, the network is not distracted by content when learning this age signal (traces).

\section{Discussion}
\label{sec:discussion}
Image age approximation can be considered as a multi-class classification problem (i.e., where the classes are defined by the temporal resolution of the exploited age traces and the available trusted images). Generally, inferences of a particular class should be based on reasonable traces. Especially in the field of image forensics, however, it is of paramount importance that decisions are based on comprehensible evidence. For example, in \cite{Mostafa25a} it is described that a Washington state superior court judge rejected the admission of AI-enhanced video exhibits because of the opaqueness of the methods \cite{CourtDecision}.

As shown in \cite{Joechl23a} the inference of a CNN trained in the context of image age approximation (described in section \ref{sec:Ahmed20b}) is most likely not based on age traces, but on image content. We also wrongly suggested in \cite{Joechl21b} (described in section \ref{sec:features}) that the SRNet utilizes unknown age traces. In fact, content bias is exploited as well (as shown in \cite{Joechl22b, Joechl23a}). These two examples and the one shown in section \ref{sec:example} demonstrate that when a standard CNN is trained naively on an age classification problem, it cannot be assumed that the inference is based on comprehensible evidence (age traces). The distraction of image content is plausible because (i) the information stored in an image largely reflects the content and only a very small fraction of it is age information, i.e., the age signal can easily be obscured by other (non-age related) correlations in the acquired data, and (ii) content biases can easily be introduced, i.e., images belonging to the same age class are usually taken in close temporal proximity (e.g., similar scenes or lighting conditions, etc.). Furthermore, as shown in subsection \ref{sec:xai_Ahmed} content bias is not limited to deep learning based methods. Thus, to mitigate content bias, new methods for image age approximation must be carefully designed. Currently, the only available and reliable methods to approximate image age are those that rely on hand-crafted features (i.e., in-field sensor defects) extracted at predefined defect positions, as proposed in \cite{Fridrich11a} and \cite{Joechl20a}.

One way to mitigate content bias could be to suppress image content, i.e., to increase the signal-to-noise ratio (age signal to image content). However, as shown in \cite{Joechl23a}, even when content suppression techniques (e.g., median filter residuals) are applied, the image content still plays a role in the decision. Another way to deal with this problem are rigorous hypothesis-driven methodologies, as advocated in \cite{Mostafa25a}. The idea is to investigate competing explanatory hypotheses. For example, in the context of image age approximation, such hypotheses could be: (H1) the model has learned to successfully predict the age classes based on the presence of in-field sensor defects; (H2) the model primarily learned features based on image content to successfully predict the age classes. A checklist for reporting standards in machine learning based science is proposed in \cite{Kapoor23a}. In context of camera identification, the `Sybil' approach is proposed in \cite{Mostafa25a}. To mitigate content bias, the idea is to divide the images of a single device according to their content (group similar images). Each group is treated as different device during training and testing, even though the images come from the same device.

Another strategy to avoid or limit content bias when training a CNN might be to apply constraints on the acquisition of training data. Currently, there is one temporal image forensics dataset, the Northumbria Temporal Image Forensics (NTIF) dataset \cite{Ahmed20a}, available. However, since regular scene images (indoor and outdoor) were taken for each time-slot, it is most likely that some content bias might also be present. This is also suggested in \cite{Joechl22b} and \cite{Joechl23a}, where the influence of image content was analyzed using the NTIF dataset.

In principle, the age signal is inherent in the artifacts and noise introduced by the image acquisition pipeline. Thus, we propose in \cite{Joechl24a} the PLUSTIF dataset, where the artifacts and noise introduced by the image acquisition pipeline are estimated based on captured calibration images (i.e., Dark Field Images (DFIs) and Bright Field Images (BFIs)) at different points in time (representing age classes). Content bias should be limited when capturing calibration images. Actually, content bias can be ruled out for DFIs, since the incident light (content) is set to zero. However, content bias cannot be ruled out when capturing BFIs, although it should be limited due to the recording procedure. The estimated artifacts and noise can then be embedded into synthetic (rendered) images. Synthetic images are completely free of any artifacts and noise introduced by the image acquisition pipeline. The idea is to embed the extracted artifacts and noise from each acquisition session (age classes) into the same set of synthetic images (e.g., the GTA5 dataset \cite{Richter16a}). Thus, truly identical images (in terms of image content) would be available per age class, only with different artifacts and noise embedded. The resulting dataset, for example, can help to develop deep learning based image age approximation methods or facilitate the discovery of unknown age traces.

In \cite{Joechl24a}, the PLUSTIF dataset is utilized as a benchmark for robustness against content bias. For this purpose, completely different types of synthetic images are used per age class (image acquisition session). This results in a highly biased dataset which is used for training. If the model is robust against content bias, the classification accuracy obtained from the model trained on the biased train-set should be similar for the biased and unbiased test-set. For the unbiased test-set, the same set of synthetic images is used for all classes. In \cite{Joechl24a}, the robustness against content bias is evaluated for: (i) the deep learning based method \cite{Ahmed20b} (described in subsection \ref{sec:Ahmed20b}), (ii) the machine learning based method where defect positions are learned \cite{Ahmed21a} (described in subsection \ref{sec:Ahmed21a}), and (iii) the machine learning method relying on features extracted at predefined defect positions \cite{Joechl20a} (described in subsection \ref{sec:Joechl20a}). As expected, the methods \cite{Ahmed20b,Ahmed21a} are prone to content bias while \cite{Joechl20a} is not.

\section{Conclusion}
\label{sec:con}
In this review, a comprehensive and critical overview of existing image age approximation methods and the most commonly used age traces is given. The open problem of content bias in deep learning based image age approximation is particularly highlighted. Based on an example, it is shown how easily a CNN can be distracted by a content bias and how problematic it is to naively train a CNN on age classes. This also emphasizes the importance of XAI methods for assessing the reliability of existing age approximation methods. Furthermore, the problem of content bias is not limited to deep learning-based methods, i.e., it is shown, that a machine learning method proposed to use in-field sensor defects for age approximation actually exploit other traces (probably due to content bias). For this reason, only methods that use dedicated age traces, where the positions are known and the behavior over time is detectable (e.g., onset of in-field sensor defects), can be recommended at current stage (i.e., the methods proposed in \cite{Fridrich11a} and \cite{Joechl20a}).

\bibliographystyle{elsarticle-num}
\bibliography{bib}
\end{document}